\documentclass[a4paper,fleqn,colorlinks,hidelinks,bookmarks=true,bookmarksopen=true]{cas-dc}

\usepackage[numbers]{natbib}
\usepackage{subfigure}
\usepackage{amsmath}
\usepackage{cases}
\usepackage{threeparttable}
\usepackage{stfloats} 
\usepackage{bbding} 
\usepackage{framed}
\usepackage{pifont} 
\usepackage{amsfonts}
\usepackage{subeqnarray}
\usepackage{longtable}
\usepackage{supertabular}
\usepackage{setspace}
\usepackage{multirow}
\usepackage{float}
\usepackage[figuresright]{rotating}
\usepackage{booktabs}
\usepackage{mathrsfs}
\usepackage{booktabs}
\usepackage{enumitem}
\usepackage{graphicx}
\makeatletter
\newcommand{\nosemic}{\renewcommand{\@endalgocfline}{\relax}}
\newcommand{\dosemic}{\renewcommand{\@endalgocfline}{\algocf@endline}}
\let\oldnl\nl
\newcommand{\nonl}{\renewcommand{\nl}{\let\nl\oldnl}}
\makeatother
\usepackage[export]{adjustbox} 
\usepackage{booktabs}
\usepackage{setspace}
\usepackage{xcolor}
\usepackage{mathrsfs}
\usepackage{amsmath}
\usepackage{array}
\usepackage{amssymb}
\usepackage{amsthm}
\usepackage{microtype}
\usepackage{url}
\usepackage{amsfonts,amssymb}
\usepackage{dsfont}
\usepackage{mathtools}
\usepackage{geometry}
\def\tsc#1{\csdef{#1}{\textsc{\lowercase{#1}}\xspace}}
\tsc{WGM}
\tsc{QE}

\begin{document}


\let\WriteBookmarks\relax
\def\floatpagepagefraction{1}
\def\textpagefraction{.001}

\shorttitle{Data-driven Power Flow Linearization -- Part I: Theory}    

\shortauthors{M. Jia, G. Hug, N. Zhang, Z. Wang, Y. Wang, C. Kang}   

\title [mode = title]{Data-driven Power Flow Linearization: Theory}  



%

\author[inst1]{Mengshuo Jia}[type=editor,      
      orcid=0000-0002-2027-5314]

\author[inst1]{Gabriela Hug}

\author[inst2]{Ning Zhang}

\author[inst3]{Zhaojian Wang}

\author[inst4]{Yi Wang}

\author[inst2]{Chongqing Kang}






\affiliation[inst1]{organization={Department of Information Technology and Electrical Engineering, ETH Zürich},
            addressline={Physikstrasse 3}, 
            postcode={8092}, 
            state={Zürich},
            country={Switzerland}}

\affiliation[inst2]{organization={Department of Electrical Engineering, Tsinghua University},
            addressline={Shuangqing Rd 30}, 
            postcode={100084}, 
            state={Beijing},
            country={China}}

\affiliation[inst3]{organization={Department of Automation, Shanghai Jiao Tong University},
            addressline={Dongchuan Rd 800}, 
            postcode={200240}, 
            state={Shanghai},
            country={China}}

\affiliation[inst4]{organization={Department of Electrical and Electronic Engineering, The University of Hong Kong},
            addressline={Pok Fu Lam}, 
            state={Hong Kong},
            country={China}}




\begin{abstract}
This two-part tutorial dives into the field of data-driven power flow linearization (DPFL), a domain gaining increased attention with the rise of data-centric methodologies. DPFL stands out for its higher approximation accuracy, wide adaptability, and better ability to implicitly incorporate power losses and the latest system attributes. This renders DPFL a potentially superior option for managing the significant fluctuations from renewable energy sources, a step towards realizing a more sustainable energy future, by translating the higher model accuracy into increased economic efficiency and less energy losses. To conduct a complete, rigorous, and deep reexamination, this tutorial first classifies all existing DPFL methods into DPFL training algorithms and supportive techniques. Their mathematical models, analytical solutions, capabilities, limitations, and generalizability are systematically examined, discussed, and summarized. In addition, this tutorial reviews existing DPFL experiments, examining the settings of test systems, the fidelity of datasets, and the comparison made among a limited number of DPFL methods. Further, this tutorial implements extensive numerical comparisons of all existing DPFL methods (40 methods in total) and four classic physics-driven approaches, focusing on their generalizability, applicability, accuracy, and computational efficiency. Through these experiments, this tutorial aims to reveal the actual performance of all the methods (including the performances exposed to data noise or outliers), therefore guiding the selection of appropriate linearization methods for researchers. Furthermore, this tutorial discusses open research questions and future directions based on the theoretical and numerical insights gained, contributing to the progression of the field of DPFL. As the first part of the tutorial, this paper mainly reexamines DPFL theories, covering all the training algorithms and supportive techniques in depth. Numerous capabilities, limitations, and aspects of generalizability, which were previously unmentioned in the literature, have been identified. (Word Count: 9966)
\end{abstract}




\maketitle

\section{Introduction}\label{sec:Intro}

Linear power flow models, fundamental to power systems, are rigorously studied and widely applied in both academic and industrial settings \cite{9914682, molzahn2019survey}, including state estimation, unit commitment, economic dispatch, grid planning, and system control \cite{23, 8_1}, unlocking trillion-dollar markets and involving every consumer worldwide. The precision and computational efficiency of linearization approaches are crucial for efficient grid operations, minimizing energy losses, and ensuring stable energy distribution, all of which are vital for a sustainable energy system increasingly dependent on renewable sources. Enhancing the accuracy and efficiency of linear power flow models exceeds mere technical improvement --- it also represents a significant step toward a more sustainable energy future, providing immense societal and financial value. 


Indeed, power flow linearization has already been intensively explored for decades in terms of how to improve accuracy and efficiency. However, the recent influx of data-centric methodologies, which affect the areas of science, engineering, technology, and society \cite{dorfler2023data}, has restored interest in the area of linearization, leading to data-driven power flow linearization (DPFL), a rapidly evolving field \cite{Powertech}. As a consequence of the widespread deployment of phasor measurement units (PMUs) \cite{19,14, 14_16, 14_18, 26_14, 30}, advanced communication infrastructure \cite{4_22}, cutting-edge analysis methods, real-time computing power, and a broad interest in data-centric methods \cite{dorfler2023data, terzija2010wide, de2010synchronized}, DPFL methods have attracted considerable attention \cite{2}. 

DPFL methods, unlike traditional physics-driven (a.k.a., model-driven) power flow linearization algorithms \cite{molzahn2019survey}, often require no prior physical knowledge/model of the power grid \cite{2}, but only the measurements of the system to train linear models \cite{19, 11}. The benefits of DPFL approaches include but are not limited to (i) usually higher approximation precision due to being assumption-free \cite{11, 2} and customizable \cite{11, 11_14, 7, 11_13} (DPFL's errors tend to be several orders of magnitude less than the errors of state-of-the-art physic-driven methods, e.g., \cite{wang2017linear, yang2016state, fan2021error}), (ii) better applicability to cases where physical parameters are unavailable\footnote{In power systems, it is frequently encountered that physical parameters are unavailable \cite{1, 29_11, 19, 27, 4, 9027950}. This is often due to factors such as varying topologies \cite{5_10, 5_11, 19_18, 29_8, 29_9, 29_10}, deviations in line parameters \cite{19_19, 29_11}, and the undisclosed control rules of privately-owned distributed energy resources \cite{27_11, 19}.}, (iii) the implicit incorporation of power losses \cite{7, 19, 20}, (iv) the integration of up-to-date system measurements \cite{19, 7_9}, and (v) the inclusion of realistic impacts such as control actions \cite{19} or human behaviors \cite{9_15}. 

Up to this point, a variety of DPFL methodologies have been developed, falling into two primary categories: classic regression-based methods and more tailored optimization methods. The classic regression-based category utilizes standard formulations from a range of regression programming models. In contrast, the tailored optimization category modifies the classic regression framework by altering the objective functions and/or constraints. Specifically, the regression-based DPFL methods contain: (i) least squares regression and its various forms \cite{1, 16, 18, 23, 13, 7, 34, 8, 2, 29, 11, 12}, (ii) partial least squares and its different versions \cite{10, 19, 20, 17}, (iii) ridge regression and its derivatives \cite{6, 24, 4}, and (iv) support vector regression and its adaptations \cite{14, 20, 34, 27, 28}. As for the tailored optimization methods, they include (i) unconstrained programming \cite{liu2023data}, (ii) linearly constrained programming \cite{9, 11}, (iii) chance-constrained programming \cite{33}, and (iv) distributionally robust chance-constrained programming \cite{3, shao2023physical}. Overall, research on DPFL has been extensive and profound to date. A preliminary, narrative review of DPFL appeared in our previous work \cite{Powertech}. 

Nevertheless, numerous critical questions remain unresolved. \textbf{Firstly}, beyond the previously mentioned DPFL methods, which directly focus on linear model training and can thus be categorized as DPFL training methods, a variety of supportive techniques have been developed and employed in DPFL analysis to tackle practical challenges. Examples include the use of coordinate transformation \cite{1, 4, 6, 9, 14, 18, 33, 27, 28} and forgetting factors \cite{2,8,18, 17} to address the inherent nonlinearity of the alternating current (AC) power flow model, and the integration of physical knowledge into the DPFL training process \cite{33,9,18,23,10} to utilize the prior information, among others. Clearly, these supportive techniques differ significantly from DPFL training algorithms and possess a degree of generalizability. A systematic analysis of these techniques would be highly beneficial for researchers in the field. Thus, it is essential to separate DPFL supportive techniques from DPFL studies for a distinct and systematic examination. However, such an analysis has not yet been conducted, leaving it largely uncertain which of these supportive techniques can be universally applied and integrated into all DPFL training algorithms, and which cannot. \textbf{Secondly}, there is a notable absence of a unified mathematical reexamination of all the DPFL training algorithms and supporting techniques, including both the mathematical models and the solution approaches. This absence hinders researchers' ability to identify and benefit from modular methods --- those that can be seamlessly combined. Additionally, this gap limits the possibility of gaining deeper methodological insights. \textbf{Thirdly}, a comprehensive and theoretical discussion of the capabilities and limitations in both DPFL training methods and supportive techniques is notably missing. This makes it challenging to determine which methods require improvement or which specific areas need enhancements. Furthermore, it reduces the chance to develop a deeper understanding of open research questions and potential future directions in this field. \textbf{Fourthly, and most critically}, there has been no comprehensive numerical comparison of all the existing DPFL training methods and supportive techniques. The actual, comparative performance of all these methods, with respect to {\color{black}generalizability, applicability, accuracy, and computational efficiency}, remains unknown. The lack of a clear understanding of the actual performance differences among existing DPFL methods could mask the problems that are not apparent from the theoretical analysis of the capabilities and limitations, obscure the judgment of researchers within the DPFL community, and complicate the selection of appropriate linearization methods for potential users from other research fields. 

In light of the above gaps, this two-part paper contributes in the following directions:
\begin{itemize}
  \item[(i)] A comprehensive review of all the existing DPFL supportive techniques is provided, revealing the generalization potential of these supportive techniques toward existing DPFL training algorithms. 
  \item[(ii)] The mathematical derivations and solutions of all existing DPFL training and supportive methods are uniformly reexamined and thoroughly discussed, providing detailed methodological insights, and an informative tutorial for interested researchers. 
  \item[(iii)] Comprehensive and theoretical investigations of the capabilities and limitations of both DPFL training methods and supportive techniques are provided. 
  \item[(iv)] A comprehensive review of existing DPFL experiments is presented, examining the adopted test scenarios, load fluctuation settings, data sources, and the considerations for data noise/outliers. The review also reveals the existing comparisons made among DPFL approaches, outlines the limitations of current experiments, and demonstrates the critical need for a comprehensive numerical comparison of all DPFL approaches. 
  \item[(v)] An exhaustive numerical simulation of over 40 linearization methods (36 existing DPFL approaches, four newly developed DPFL methods, and four classic PPFL algorithms) is conducted. A detailed comparative analysis of these 44 methods is presented, discussing their generalizability, applicability, accuracy, and computational efficiency, thereby clarifying the actual performance of all the evaluated approaches.
  \item[(vi)] An in-depth discussion regarding the open research questions and potential future directions in the DPFL field is provided. 
\end{itemize}

The first part of this two-part paper primarily reexamines and analytically discusses the theory of DPFL, encompassing both the training algorithms and supportive techniques, along with an evaluation of their capabilities and limitations. The second part is focused on reviewing existing experiments and conducting numerical simulations of all DPFL methods, leading to an exhaustive evaluation and comparison of their actual performances, as well as a detailed elaboration of open questions and future directions in the field. 

This paper, as the first part, is structured as follows: Section II discusses the DPFL training algorithms, Section III reviews the DPFL supportive techniques, and Section IV concludes the paper. For the second part, see \cite{partII}.

\paragraph{Notation}~\\

 We use $\widetilde{\boldsymbol{x}} \in \mathbb{R}^{\widetilde{N}_x \times 1}$ to denote the vector consisting of $\widetilde{N}_x$ independent variables, including active power injections of PQ and PV nodes, voltage magnitudes of PV nodes, and the voltage phase angle of the slack node. We further define $\boldsymbol{x} = [1, \widetilde{\boldsymbol{x}}^{\top}]^{\top} \in \mathbb{R}^{N_x \times 1}$, where ``$1$'' corresponds to the constant term in the linear power flow model described later, and $N_x = \widetilde{N}_x + 1$. Furthermore, we use $\boldsymbol{y} \in \mathbb{R}^{N_y \times 1}$ to refer to the vector that collects $N_y$ dependent variables, including voltage magnitudes of PQ nodes, voltage phase angles of PQ and PV nodes, active/reactive power injections of the slack node, line flows, and line power losses. Moreover, we use $\boldsymbol{x}_i \in \mathbb{R}^{N_x \times 1}$ and $\boldsymbol{y}_i \in \mathbb{R}^{N_y \times 1}$ to represent the $i$-th measurements of $\boldsymbol{x}$ and $\boldsymbol{y}$, respectively. Correspondingly, $x_{ij}$ refers to the $j$-th element of $\boldsymbol{x}_i$; the same applies to $y_{ij}$. The datasets of $\boldsymbol{x}$ and $\boldsymbol{y}$ are expressed by $\boldsymbol{X} = [\boldsymbol{x}_1, \ \cdots, \boldsymbol{x}_{N_s} ]^{\top} \in \mathbb{R}^{N_s \times N_x}$ and $\boldsymbol{Y} = [\boldsymbol{y}_1, \ \cdots, \boldsymbol{y}_{N_s}]^{\top} \in \mathbb{R}^{N_s \times N_y}$, respectively; each of them contains $N_s$ measurements. Additionally, we use $v$ and $\theta$ to respectively express voltage magnitude and phase angle. Particularly, $v_i$ denotes the voltage magnitude at bus $i$, $\theta_{ij}$ refers to the angle difference between buses $i$ and $j$, and $P_{ij}$ ($Q_{ij}$) represents the active (reactive) branch flow of the line connecting buses $i$ and $j$. Meanwhile, the active and reactive power injections of bus $i$ are described by:
\begin{align}
  P_i & = v_i\sum\nolimits_{j=1}^{N_b} v_j\left(G_{ij}\cos\theta_{ij}+B_{ij}\sin\theta_{ij} \right) \label{eq:Pi}  \\
  Q_i & = v_i\sum\nolimits_{j=1}^{N_b} v_j\left(G_{ij}\sin\theta_{ij}-B_{ij}\cos\theta_{ij} \right) \label{eq:Qi}
\end{align}

where $G_{ij}$ and $B_{ij}$ refer to the real and imaginary parts of entry $(i, j)$ in the nodal admittance matrix, respectively, and $N_b$ represents the number of buses. Note that for the sake of brevity, in the following, we use the term ``voltage'' to particularly indicate the voltage magnitude while using ``angle'' to denote the voltage phase angle. 

\paragraph{Problem Formulation}~\\

DPFL aims to identify an optimal linear relationship between $\boldsymbol{x}$ and $\boldsymbol{y}$ {\color{black} (a.k.a., predictor and response, respectively)} using $\boldsymbol{X}$ and $\boldsymbol{Y}$. The relationship is parameterized by a  coefficient matrix, generally denoted by $\boldsymbol{\beta}$. Various forms of linear relationships have been used in existing DPFL studies. We denote them by
\begin{align}
  \text{\textbf{Model 1:}} \quad & \boldsymbol{y} = \boldsymbol{\beta}^{\top}\boldsymbol{x} \label{eq:linear_pure} \\
  \text{\textbf{Model 2:}} \quad & \boldsymbol{y} = \boldsymbol{\beta}(k)^{\top}\boldsymbol{x}, \ k = 1 \ \cdots K\label{eq:linear_K} \\
  \text{\textbf{Model 3:}} \quad & \boldsymbol{y} = \boldsymbol{\beta}_{\phi}^{\top}\phi(\boldsymbol{x}) \label{eq:linear_map} 
\end{align}
\textbf{Model 1} is a single linear model with  $\boldsymbol{\beta} \in \mathbb{R}^{N_x \times N_y}$ being the coefficient matrix (we denote $\boldsymbol{\beta}_j \in \mathbb{R}^{N_x \times 1}$ as the $j$-th column of $\boldsymbol{\beta}$ hereafter). \textbf{Model 2} is a piecewise linear model, where $\boldsymbol{\beta}(k) \in \mathbb{R}^{N_x \times N_y}$ is the coefficient matrix for segment $k$, and $K$ denotes the number of total segments. \textbf{Model 3}, parameterized by $\boldsymbol{\beta}_{\phi} \in \mathbb{R}^{N_{\phi} \times N_y}$, describes the linear relationship between $\boldsymbol{y}$ and $\phi(\boldsymbol{x})$, where $\phi \colon \mathbb{R}^{N_x} \to \mathbb{R}^{N_{\phi}}$ is a mapping function. Note that the coefficients in the above models are estimated via data mining. For distinction, we use $\wedge$ to highlight the estimation result, e.g., $\hat{\boldsymbol{\beta}}$ represents the estimated coefficient matrix of \textbf{Model 1}, and $\hat{\boldsymbol{y}}$ denotes the estimated (or say predicted) value of $\boldsymbol{y}$. 
Given that \textbf{Model 1} is the most widely used model compared to the other two, we treat this model as a default in the following sections, unless otherwise specified. 

\vspace{0.3cm}

\noindent \textbf{Remark}: \textit{As one of the contributions, this paper thoroughly analyzes the capabilities and limitations of all DPFL methods. It is crucial to acknowledge that no method is flawless. Every approach, either a DPFL training algorithm or a supportive technique, has its notable capabilities and significance. While these methods may exhibit various limitations, it is essential to understand that the weakness analysis in this paper is not a criticism, but merely a part of a comprehensive evaluation. The identification of limitations is also approached from the standpoint of an idealized ``perfect method,'' a concept that, in reality, does not exist. This rigorous examination aims to provide a complete and balanced assessment.}

\section{Training Algorithm}\label{sec:alg} 
\begin{figure*}[]
	\centering 
	\includegraphics[width=7.3in]{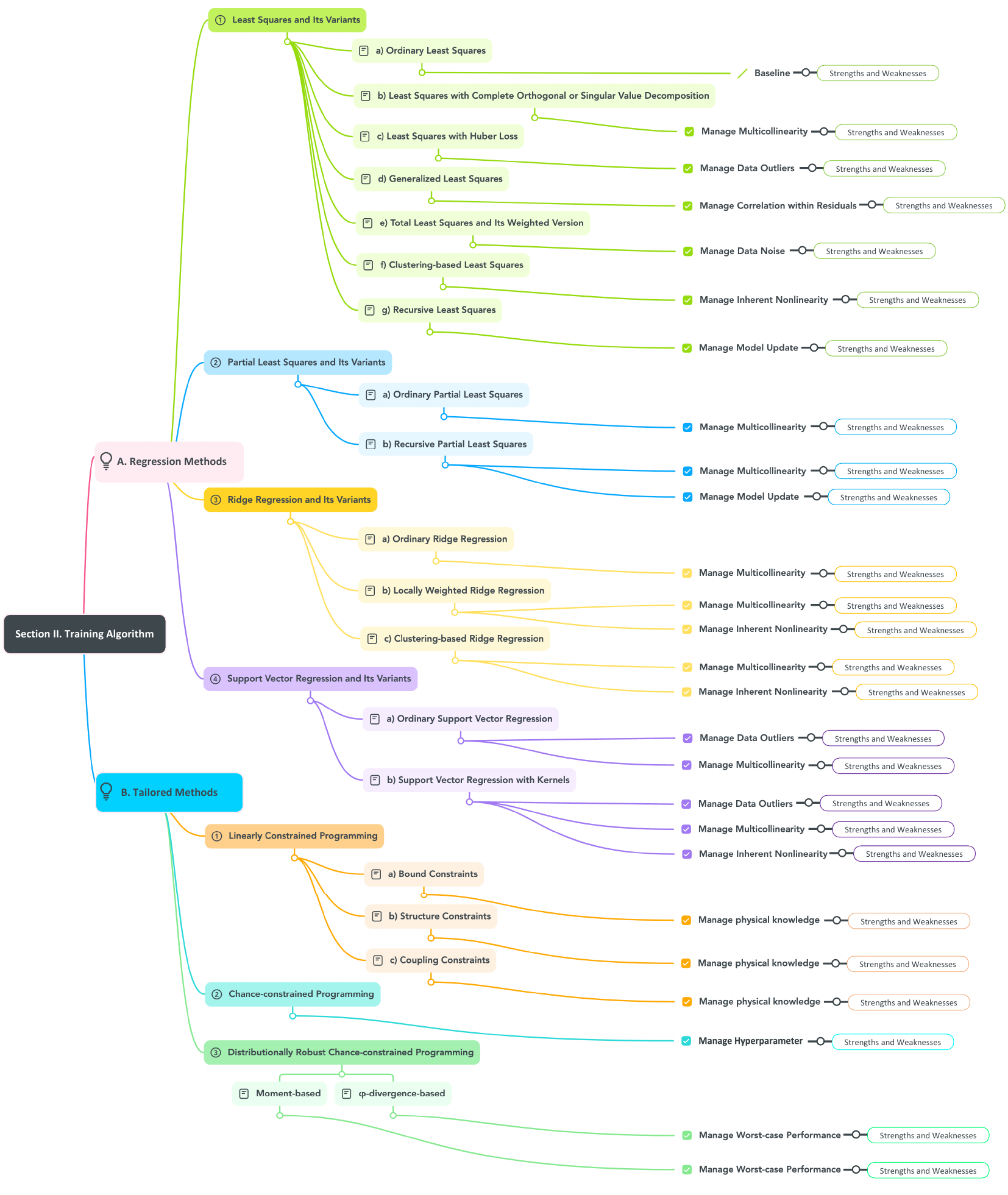} 
	\vspace{-0.4cm}
	\caption{Graphical mirror of the outline of Section II, also explaining the classifications, interconnections, and initial motivations (i.e., ``Manage ...'') of existing DPFL training algorithms. }
	\label{Fig:TrainAlgo} 
\end{figure*}

{\color{black} 
This section reexamines DPFL training algorithms, categorizing them into two primary types: regression and tailored algorithms. These algorithms, together with their capabilities and limitations, are further analyzed below. For a clearer, visual representation of these algorithms, including their classifications, interconnections, and initial motivations, please see Fig. \ref{Fig:TrainAlgo}, where a detailed graphical logic tree is provided, also mirroring the structure of this section. 

}

\subsection{Regression Methods}\label{sec:regression}
Regression-based DPFL training methods include the least squares regression, partial least squares regression, ridge regression, support vector regression, and their variants. In the following, these different approaches are revisited.

\subsubsection{Least Squares and Its Variants}\label{sec:OLS} 

{\color{black}We start with the ordinary least squares method, and then move to its various variants. Each of the variants addresses specific shortcomings of the ordinary least squares method.}

\paragraph{Ordinary Least Squares}~\\

This method, used in the DPFL work from \cite{18}, adopts the following regression model
\begin{align}
  \boldsymbol{Y} &  = \boldsymbol{X}\boldsymbol{\beta} + \boldsymbol{\varepsilon} \label{eq:LinMod}
\end{align}
where $\boldsymbol{\varepsilon} \in \mathbb{R}^{N_s \times N_y}$ collects the residuals for observations. The ordinary least squares approach finds the coefficient matrix $\boldsymbol{\beta}$ by minimizing the sum of squared residuals:
\begin{align}
  \min \limits_{\boldsymbol{\beta}} \   \Vert \boldsymbol{Y}- \boldsymbol{X}\boldsymbol{\beta} \Vert_{F}^2 \label{eq:RSS}
\end{align}
where $\Vert \cdot \Vert_F$ is the Frobenius norm (a.k.a., Hilbert–Schmidt norm). The solution to the above unconstrained programming problem, i.e., the estimation of $\boldsymbol{\beta}$, can be expressed by \cite{27}
\begin{align}
  \hat{\boldsymbol{\beta}} = \left(\boldsymbol{X}^{\top} \boldsymbol{X} \right)^{-1} \boldsymbol{X}^{\top}\boldsymbol{Y} \label{eq:beta_esti}
\end{align}

\noindent \textit{\textbf{Capabilities}}: This method is easy to implement and provides a clear, interpretable solution.

\noindent \textit{\textbf{Limitations}}: This method has the following limitations, labeled \textbf{L1}, \textbf{L2}, etc., for easy reference and comparison with other DPFL methods.
\begin{itemize}
  \item[\textbf{L1}] The method requires $\boldsymbol{X}$ to be of full column rank. However, in power systems, $\boldsymbol{X}$ often has highly correlated columns due to similar voltage patterns of connected nodes \cite{24_6,4,14} and active power injections \cite{19}. This multicollinearity \cite{1,4,24_6}, also known as data coherence \cite{26} or isotropic dispensation \cite{24}, negatively affects the method's performance, causing errors, chaos, sample-dependent performance, and overfitting \cite{16,1}. If $\boldsymbol{X}$ is not full rank, $\boldsymbol{X}^{\top} \boldsymbol{X}$ is non-invertible, making $\hat{\boldsymbol{\beta}}$ in \eqref{eq:beta_esti} nonexistent.
  \item[\textbf{L2}] The method is sensitive to data outliers, since the squared-error loss function in \eqref{eq:RSS} emphasizes large residuals, heavily influencing training and degrading model performance.
  \item[\textbf{L3}] The method assumes $\mathbb{V}ar(\boldsymbol{\varepsilon}|\boldsymbol{X}) = \sigma_{\varepsilon}^2 \boldsymbol{I}$, implying homoskedasticity and no autocorrelation ( $\sigma_{\varepsilon}^2$ is a constant, $\boldsymbol{I}$ is the identify matrix, and $\mathbb{V}ar(\cdot)$ denotes the variance function). However, in power systems, voltage magnitudes and time-series data in $\boldsymbol{Y}$ are often correlated \cite{11}, violating this assumption.
  \item[\textbf{L4}] The DPFL model needs continuous updates with new observations to stay current. While the ordinary least squares algorithm can incorporate new data, the process becomes increasingly demanding as the sizes of $\boldsymbol{X}$ and $\boldsymbol{Y}$ grow, especially in large systems.
  \item[\textbf{L5}] This method does not account for the inherent nonlinearity in the AC power flow model.
  \item[\textbf{L6}] The method struggles with noisy data, as such data often violate the assumptions of homoskedasticity and no autocorrelation, yet power systems' measurements are frequently noisy.
\end{itemize}

\paragraph{Least Squares with Complete Orthogonal (or Singular Value) Decomposition}~\\

To deal with the singularity caused by multicollinearity, the complete orthogonal decomposition has been used in DPFL studies \cite{7, 13}. This decomposition, as an extension of the QR decomposition \cite{7_28}, factorizes $\boldsymbol{X}^{\top} \boldsymbol{X}$ into
\begin{align}
  \boldsymbol{X}^{\top} \boldsymbol{X} =  \widetilde{\boldsymbol{Q}} \left[ \begin{array}{cc}
    \widetilde{\boldsymbol{R}} & \boldsymbol{0}  \\
    \boldsymbol{0} & \boldsymbol{0} \\
    \end{array} \right] \boldsymbol{Z}^{\top} \label{eq:COD}
\end{align}
regardless of the rank of $\boldsymbol{X}^{\top} \boldsymbol{X}$, where $\widetilde{\boldsymbol{Q}} \in \mathbb{R}^{N_x \times N_x} $ and $\boldsymbol{Z} \in \mathbb{R}^{N_x \times N_x} $ are both orthogonal matrices, and $\widetilde{\boldsymbol{R}} \in \mathbb{R}^{N_{x_r} \times N_{x_r}} $ is an upper-triangular matrix. Substituting \eqref{eq:COD} into \eqref{eq:beta_esti} yields 
\begin{align}
  \hat{\boldsymbol{\beta}} = \boldsymbol{Z} \left[ \begin{array}{cc}
    \widetilde{\boldsymbol{R}}^{-1} & \!\!\! \boldsymbol{0}  \\
    \boldsymbol{0} \quad\quad & \!\!\! \boldsymbol{0} \\
    \end{array} \right] \widetilde{\boldsymbol{Q}}^{\top} \boldsymbol{X}^{\top} \boldsymbol{Y} \label{eq:beta_esti_COD}
\end{align}

Note that the singular value decomposition (SVD) can also address the singularity issue \cite{18}. Similar to the previously described method, SVD is applied to $\boldsymbol{X}^{\top} \boldsymbol{X}$, and the decomposition result is then used in \eqref{eq:beta_esti} to estimate $\boldsymbol{\beta}$. Details are omitted for brevity.

{\color{black}
\noindent \textit{\textbf{Capabilities}}: \textit{Firstly}, the least squares with the complete orthogonal (or singular value) decomposition method can generally handle the multicollinearity issue in $\boldsymbol{X}$. \textit{Secondly}, it can speed up the regression progress, because $\widetilde{\boldsymbol{R}}$ is an upper-triangular matrix with fewer dimensions, rendering $\widetilde{\boldsymbol{R}}^{-1}$ in \eqref{eq:beta_esti_COD} easier to compute. 

\noindent \textit{\textbf{Limitations}}: \textit{Firstly}, the least squares with singular value decomposition method cannot guarantee the inversion of $\boldsymbol{X}^{\top} \boldsymbol{X}$ during DPFL model training, especially when $\boldsymbol{X}^{\top} \boldsymbol{X}$ has zero singular values. \textit{Secondly}, this method also shares the limitations of the ordinary least squares method, including \textbf{L2}, \textbf{L3}, \textbf{L4}, \textbf{L5}, and \textbf{L6}.

}

\paragraph{Least Squares with Huber Loss}~\\

To improve robustness to data outliers, the Huber loss function has been applied to DPFL \cite{8, 2}. The Huber loss function replaces the squared loss of outliers with their absolute loss, reducing the impact of bad data. Specifically, the Huber loss function $H_i(\cdot)$ transforms \eqref{eq:RSS} into:
\begin{align}
  \min \limits_{\boldsymbol{\beta}} \   \sum\nolimits_{i=1}^{N_s}\sum\nolimits_{j=1}^{N_y} H_i(\boldsymbol{\beta}_j) \label{eq:hub_obj}
\end{align}
with
\begin{equation}
  H_i(\boldsymbol{\beta}_j) = \left\{
		\begin{split}
		& \varepsilon_{ij}^2,  &  \Vert\varepsilon_{ij}\Vert_1 \leq \delta^{HUB} &  \\
		& \delta^{HUB}\left( 2 \Vert\varepsilon_{ij}\Vert_1 - \delta^{HUB} \right) , & \Vert \varepsilon_{ij} \Vert_1 > \delta^{HUB} & \\
		\end{split} \right. \notag
\end{equation}
where $\Vert \cdot \Vert_1$ denotes the $\ell^1$ norm, $\varepsilon_{ij} = y_{ij}-\boldsymbol{x}_{i}^{\top}\boldsymbol{\beta}_j$, and $\delta^{HUB} \in \mathbb{R}$ is a preset threshold. This threshold is the key to distinguishing between inliers and outliers. The tuning of $\delta^{HUB}$ is thus crucial and should be carried out using systematic approaches, such as the bisectional search \cite{8_35}. Note that the above regression problem can be solved either directly or by converting \eqref{eq:hub_obj} into an equivalent convex programming model \cite{8, 8_34}. 

{\color{black}

\noindent \textit{\textbf{Capabilities}}: This method is more robust against data outliers and less sensitive to multicollinearity compared to the ordinary least squares approach.

\noindent \textit{\textbf{Limitations}}: \textit{Firstly}, this method lacks closed-form solutions, and solving the optimization problem in \eqref{eq:hub_obj} can be time-intensive, especially for large power flow models, due to the numerous entries in $\boldsymbol{\beta}$ treated as decision variables. This complexity makes the problem challenging to solve, despite its convex nature. \textit{Secondly}, this method relies heavily on tuning $\delta^{HUB}$, which typically requires solving \eqref{eq:hub_obj} repeatedly, leading to substantial computational demands. \textit{Thirdly}, there are no guarantees regarding noise efficacy, as it depends on the noise level. \textit{Lastly}, this method also shares some common limitations, including \textbf{L3}, \textbf{L4}, and \textbf{L5}.

}

\paragraph{Generalized Least Squares}~\\

The generalized least squares method takes the conditional correlation within $\boldsymbol{\varepsilon}$ into account, in order to find the best linear unbiased estimation of $\boldsymbol{\beta}$. This method has been utilized in \cite{29} for DPFL. Specifically, let us denote the ground truth of the conditional variance-covariance matrix of residuals as
\begin{align}
  \mathbb{V}ar(\boldsymbol{\varepsilon}|\boldsymbol{X}) = \sigma_{\varepsilon}^2 \boldsymbol{\Omega} \label{eq:GLS_assume}
\end{align}
where $\boldsymbol{\Omega} \in \mathbb{R}^{N_s \times N_s}$ is a positive definite, symmetric, but not necessarily diagonal matrix. The diagonal elements of $\boldsymbol{\Omega}$ can also differ from each other. Once $\boldsymbol{\Omega}$ is known, the solution of the generalized least squares method can be expressed by
\begin{align}
  \hat{\boldsymbol{\beta}} = \left(\boldsymbol{X}^{\top} \boldsymbol{\Omega}^{-1} \boldsymbol{X} \right)^{-1} \boldsymbol{X}^{\top}\boldsymbol{\Omega}^{-1} \boldsymbol{Y} \label{eq:beta_esti_GLS}
\end{align} 
which, according to the Gauss-Markhov Theorem, is the best linear unbiased estimation of $\boldsymbol{\beta}$ under the heteroskedastic and autocorrelated circumstance \cite{davidson2004econometric}. Noticeably, if $\boldsymbol{\Omega}$ degenerates into an identity matrix, \eqref{eq:beta_esti_GLS} simplifies to \eqref{eq:beta_esti}. 

{\color{black}
\noindent \textit{\textbf{Capabilities}}: This method can produce the best linear unbiased estimation of $\boldsymbol{\beta}$ if the ground truth of the conditional variance-covariance matrix of residuals is known beforehand.

\noindent \textit{\textbf{Limitations}}: \textit{Firstly}, the true conditional variance-covariance matrix $\boldsymbol{\Omega}$ of $\boldsymbol{\varepsilon}$ is typically unknown and difficult to estimate in real-world applications. \textit{Secondly}, strong assumptions are usually needed to estimate $\boldsymbol{\Omega}$. For example, \cite{29} assumes that only measurement errors contribute to the regression residuals, disregarding the errors in the regression itself; it also assumes measurement errors are present only in the dependent variables, and that these errors follow a Gaussian distribution with zero mean. These assumptions may not hold in power systems. If regression errors and measurement errors of independent variables are considered, a more sophisticated analysis is required to find the true value of $\boldsymbol{\Omega}$. \textit{Thirdly}, if multicollinearity causes $\boldsymbol{X}$ to lack full column rank, introducing $\boldsymbol{\Omega}$ may not guarantee the invertibility of $\boldsymbol{X}^{\top} \boldsymbol{\Omega}^{-1} \boldsymbol{X}$. Thus, this method still faces challenges with multicollinearity, a frequent issue in power systems. \textit{Lastly}, this method shares some common shortcomings, including \textbf{L2}, \textbf{L4}, \textbf{L5}, and \textbf{L6}.
}

\paragraph{Total Least Squares and Its Weighted Version}~\\

To handle the noise issue, the total least squares method has been used in  DPFL \cite{11}. For illustration, let us first interpret the ordinary least squares method from a different perspective. The ordinary least squares method, unlike the total least squares, assumes that only the observation of $\boldsymbol{y}$, i.e., $\boldsymbol{Y}$, contains observational noise, whereas $\boldsymbol{X}$ does not, i.e.,
\begin{align}
    \boldsymbol{Y}=\boldsymbol{Y}_0+\boldsymbol{\varepsilon}_y, \ \boldsymbol{X}=\boldsymbol{X}_0 
\end{align}
where $\boldsymbol{Y}_0 \in \mathbb{R}^{N_s \times N_y} $ and $\boldsymbol{X}_0 \in \mathbb{R}^{N_s \times N_x}$ denote the ground truth realizations of $\boldsymbol{y}$ and $\boldsymbol{x}$, respectively, and $\boldsymbol{\varepsilon}_y \in \mathbb{R}^{N_s \times N_y}$ refers to the noise when observing $\boldsymbol{y}$. The true linear relationship between $\boldsymbol{x}$ and $\boldsymbol{y}$ should be
\begin{align}
  \boldsymbol{Y}_0 = \boldsymbol{X}_0 \boldsymbol{\beta} \label{eq:real_lin}
\end{align}
However, given that only $\boldsymbol{Y}$ instead of $\boldsymbol{Y}_0$ is available, in order to estimate $\boldsymbol{\beta}$ that solves \eqref{eq:real_lin}, the ordinary least squares method introduces a perturbation $\Delta\boldsymbol{Y}$ to compensate for the noise $\boldsymbol{\varepsilon}_y$, i.e., the target is to realize 
\begin{align}
   \boldsymbol{Y} + \Delta \boldsymbol{Y} = \boldsymbol{X}_0\boldsymbol{\beta} 
\end{align}
Geometrically, $\Delta\boldsymbol{Y}$ represents the vertical distance between the data point in $\boldsymbol{Y}$ and the fitting hyperplane $\boldsymbol{X}_0\boldsymbol{\beta}$. The obtained fitting hyperplane should ensure the distance $\Delta\boldsymbol{Y}$ to be minimal. Hence, the programming model of the ordinary least squares method is formulated as
\begin{equation}
  \begin{aligned}
    \min_{\boldsymbol{\beta}} \quad  & \Vert \Delta \boldsymbol{Y} \Vert_{F}^2  \label{eq:LS_ori} \\
  \textrm{s.t.} \quad  & \Delta \boldsymbol{Y} = \boldsymbol{X}_0\boldsymbol{\beta} - \boldsymbol{Y} \\
  \end{aligned} 
\end{equation}
which is equivalent to \eqref{eq:RSS}. The total least squares approach, on the other hand, treats $\boldsymbol{Y}$ and $\boldsymbol{X}$ symmetrically, i.e., $\boldsymbol{X}$ is also noisy. This is reasonable, as $\boldsymbol{Y}$ and $\boldsymbol{X}$ are both measured in reality \cite{overview}. Correspondingly, $\boldsymbol{X} = \boldsymbol{X}_0 + \boldsymbol{\varepsilon}_x$ holds, where $\boldsymbol{\varepsilon}_x \in \mathbb{R}^{N_s \times N_x}$ is the noise when observing $\boldsymbol{x}$. Again, to compensate for $\boldsymbol{\varepsilon}_x$, a perturbation of $\boldsymbol{X}$, namely $\Delta \boldsymbol{X} \in \mathbb{R}^{N_s \times N_x}$, is integrated into the regression model, leading to the following programming problem:
\begin{equation}
  \begin{aligned}
    \min_{\boldsymbol{\beta}} \quad  & \Vert \left[\Delta \boldsymbol{X} \ \  \Delta \boldsymbol{Y}\right] \Vert_{F}^2  \label{eq:TLS} \\
  \textrm{s.t.} \quad  & \boldsymbol{Y} + \Delta \boldsymbol{Y} = \left(\boldsymbol{X} + \Delta \boldsymbol{X}\right) \boldsymbol{\beta} \\
  \end{aligned} 
\end{equation}
which minimizes the orthogonal distance  between the data point in $\left[ \boldsymbol{X} \  \   \boldsymbol{Y}\right]$ and the fitting hyperplane.  

The classic total least squares model in \eqref{eq:TLS} can be solved using the singular value decomposition \cite{overview}. Specifically, let us first denote the singular value decomposition of matrix $\left[ \boldsymbol{X} \  \   \boldsymbol{Y}\right]$ as
\begin{align}
\left[ \boldsymbol{X} \ \    \boldsymbol{Y}\right] = \boldsymbol{U}\boldsymbol{\Sigma}\boldsymbol{V}^{\top}
\end{align}
where $\boldsymbol{V} \in \mathbb{R}^{(N_x + N_y) \times (N_x + N_y) } $ is partitioned as
\begin{align}
  \boldsymbol{V} = \left[ \begin{array}{cc}
    \boldsymbol{V}_{xx} & \boldsymbol{V}_{xy}  \\
    \boldsymbol{V}_{yx} & \boldsymbol{V}_{yy} \\
    \end{array} \right]  \notag
\end{align}
with $\boldsymbol{V}_{xy} \in \mathbb{R}^{N_x \times N_y}$. Accordingly, the solution is given by
\begin{align}
\hat{\boldsymbol{\beta}} = - \boldsymbol{V}_{xy}\boldsymbol{V}_{yy}^{-1} 
\end{align}
if and only if $\boldsymbol{V}_{yy}$ is non-singular. Particularly, when $N_y=1$ and $\boldsymbol{X}$ is full rank, the above solution can be further given by
\begin{align}
\hat{\boldsymbol{\beta}} = \left[\boldsymbol{X}^{\top} \boldsymbol{X} -  (\lambda^{TLS})^2\boldsymbol{I} \right]^{-1} \boldsymbol{X}^{\top}\boldsymbol{Y} 
\end{align}
where $\boldsymbol{I}$ is an identity matrix with appropriate dimension, and $\lambda^{TLS}$ is the smallest singular value of $\left[ \boldsymbol{X}\ \   \boldsymbol{Y}\right]$, which is also seen as an estimation for the standard deviation of the noise \cite{overview}. Compared to the solution given in \eqref{eq:beta_esti}, the solution provided here removes the bias caused by the noise by subtracting the noise's variance matrix, i.e., $(\lambda^{TLS})^2\boldsymbol{I}$, from $\boldsymbol{X}^{\top} \boldsymbol{X}$. This is an intuitive explanation of why the total least squares can better handle data noise.

Note that the total least squares method assumes that the standard deviations of measurement noise from different devices are equal, which might be impractical in power systems. Hence, the weighted total least squares approach has been further used for DPFL analysis \cite{11}. The introduction of weights can treat the noises from various measurement devices unequally, in order to account for the difference in the standard deviations of noises. Mathematically, the weighted total least squares model is given by:
\begin{equation}
  \begin{aligned}
    \min_{\boldsymbol{\beta}} \quad  & \Vert \left[\Delta \boldsymbol{X} \ \  \Delta \boldsymbol{Y}\right] \Vert_{\Sigma}^2  \label{eq:WTLS} \\
  \textrm{s.t.} \quad  & \boldsymbol{Y} + \Delta \boldsymbol{Y} = \left(\boldsymbol{X} + \Delta \boldsymbol{X}\right) \boldsymbol{\beta} \\
  \end{aligned} 
\end{equation}
where the weighted matrix norm $\Vert\cdot \Vert_{\Sigma}^2$ refers to
\begin{align}
\Vert \boldsymbol{H} \Vert_{\Sigma}^2 = \rm{vec}(\boldsymbol{H})^{\top}\boldsymbol{\Sigma}^{-1}\rm{vec}(\boldsymbol{H})
\end{align}
for any given matrix $\boldsymbol{H}$; operator $\rm{vec}(\cdot)$ reshapes a matrix into a vector; $\boldsymbol{\Sigma}$ is a diagonal matrix, whose diagonal elements are usually assumed to be the variances of the corresponding noises.

{\color{black}
\noindent \textit{\textbf{Capabilities}}: \textit{Firstly}, this method offers the advantage of mitigating the impact of measurement noise. \textit{Secondly}, the weighed version has the capability to treat noises from different devices in a non-uniform manner. 

\noindent \textit{\textbf{Limitations}}: \textit{Firstly}, the standard total least squares method applies equal treatment to noise from various devices. Its weighted variant allows for unequal treatment but results in a non-convex problem lacking a closed-form solution and may yield local optima \cite{overview, 11}. \textit{Secondly}, solution methods for the weighted total least squares problem are heuristic and based on strong assumptions. For example, \cite{11} transforms the problem into approximated linearly constrained quadratic problems, assuming identical structures for $\boldsymbol{\beta}$ and the AC power flow model's Jacobian matrix, symmetry of the Jacobian matrix, and uniform voltage magnitudes, which may not always hold. \textit{Thirdly}, implementing the weighted total least squares method is challenging, as shown by the impractical step of noise removal from measurements in \cite{11}. \textit{Fourthly}, solving the approximated weighted total least squares problem is computationally intensive for large power systems due to the many decision variables in $\boldsymbol{\beta}$. \textit{Fifthly}, in the standard total least squares method, when multicollinearity causes $\boldsymbol{X}$ to be rank-deficient, subtracting $(\lambda^{TLS})^2\boldsymbol{I}$ from $\boldsymbol{X}^{\top} \boldsymbol{X}$ generally does not result in an invertible matrix. The standard method thus still faces multicollinearity challenges, though the weighted method might be less sensitive as it does not require inverting $[\boldsymbol{X}^{\top} \boldsymbol{X} -  (\lambda^{TLS})^2\boldsymbol{I}]$. \textit{Lastly}, both the original and weighted total least squares methods share some common drawbacks, namely \textbf{L3}, \textbf{L4}, and \textbf{L5}.

}

\paragraph{Clustering-based Least Squares}\label{sec:cluster_OLS}~\\

To address the nonlinearity of the AC power flow model, researchers propose using separate linear models for different operating modes. Clustering algorithms can distinguish between these modes. A clustering-based least squares method is thus developed in \cite{12} for DPFL, combining clustering with the ordinary least squares method to create a piecewise linear model, \textbf{Model 2}. Specifically, $\boldsymbol{X}$ is divided into $K$ clusters using a classic algorithm such as K-means or Gaussian mixture model \cite{12}. The datasets from the $k$-th cluster are then input into the ordinary least squares, generating the regression coefficient for the $k$-th segment, i.e.,
\begin{align}
\hat{\boldsymbol{\beta}}(k) = \left[\boldsymbol{X}(k)^{\top} \boldsymbol{X}(k)  \right]^{-1} \boldsymbol{X}(k)^{\top} \boldsymbol{Y}(k), \ \forall k
\end{align}

The application of the resulting piecewise model is straightforward: for a given input $\boldsymbol{x}$, one first identifies the cluster that $\boldsymbol{x}$ belongs to using 
\begin{align} 
k = arg \min \limits_{j} \Vert \boldsymbol{x} - \boldsymbol{\mu}(j)\Vert_F^2 
\end{align}
where $\boldsymbol{\mu}(j)$ is the centroid of cluster $j$. Then, substituting $\boldsymbol{x}$ into $\hat{\boldsymbol{y}} = \hat{\boldsymbol{\beta}}(k)^{\top}\boldsymbol{x}$ generates the prediction of $\boldsymbol{y}$.

{\color{black}
\noindent \textit{\textbf{Capabilities}}: This method addresses the AC power flow model's nonlinearity by integrating clustering into training, potentially increasing the DPFL model's accuracy.

\noindent \textit{\textbf{Limitations}}: \textit{Firstly}, integrating the piecewise linear model into decision-making frameworks, like optimal power flow problems, can complicate the framework and introduce extra integer variables, increasing problem-solving difficulty. \textit{Secondly}, tuning the number of clusters adds computational burden. \textit{Thirdly}, in large power systems with many independent variables, numerous clusters can lead to underdetermined linear systems with fewer than $N_y$ observations per cluster, yielding invalid results. \textit{Lastly}, this method shares most limitations of the ordinary least squares method, including \textbf{L1}, \textbf{L2}, \textbf{L3}, \textbf{L4}, and \textbf{L6}.
}

\paragraph{Recursive Least Squares}\label{sec:RLS}~\\

For a more computationally efficient update process, the recursive least squares algorithm \cite{7084175} offers an alternative and has been used for online updating of DPFL models \cite{liu2023data}.

Recursive least squares is an adaptive algorithm that updates regression coefficients as new data becomes available. Let $\boldsymbol{X}[t] \in \mathbb{R}^{t \times N_x}$ and $\boldsymbol{Y}[t] \in \mathbb{R}^{t \times N_y}$ represent measurements of $\boldsymbol{x}$ and $\boldsymbol{y}$ from time step 1 to $t$. Suppose $\boldsymbol{\beta}[t]$ has been established by ordinary least squares using by $\boldsymbol{X}[t]$ and $\boldsymbol{Y}[t]$. When new measurements at time step $t+1$, $\boldsymbol{x}_{t+1}$ and $\boldsymbol{y}_{t+1}$, are available, the recursive least squares algorithm updates the coefficient vector $\boldsymbol{\beta}$ via:
\begin{align}
\boldsymbol{\beta}[t+1] = \boldsymbol{\beta}[t] + \boldsymbol{K}[t+1] (\boldsymbol{y}_{t+1} - \boldsymbol{\beta}[t]^{\top} \boldsymbol{x}_{t+1}) 
\end{align}
where \( \boldsymbol{K}[t+1] \in  \mathbb{R}^{N_x \times 1} \) is the gain vector, computed as:
\begin{align} 
\boldsymbol{K}[t+1] = \frac{\boldsymbol{P}[t] \boldsymbol{x}_{t+1}}{\kappa + \boldsymbol{x}_{t+1}^{\top} \boldsymbol{P}[t] \boldsymbol{x}_{t+1}} 
\end{align}
Here, \( \boldsymbol{P}[t] \in \mathbb{R}^{N_x \times N_x} \) is the inverse of the covariance matrix of \( \boldsymbol{X}[t] \), and can be updated by
\begin{align}  
\boldsymbol{P}[t+1] = \frac{1}{\kappa} \left( \boldsymbol{P}[t] - \boldsymbol{K}[t+1] \boldsymbol{x}_{t+1}^{\top} \boldsymbol{P}[t] \right) 
\end{align}
Parameter \(\kappa \in \mathbb{R}\) is a forgetting factor between 0 and 1. This factor is useful in non-stationary environments where data may change over time. By giving more weight to recent observations (smaller $\kappa$), the recursive least squares algorithm adapts quickly to changes. Note that: (i) When $\kappa=1$, recursive least squares is equivalent to ordinary least squares when updating with new data; (ii) $\kappa$ can be time-variant; see \cite{7084175} for examples.

\noindent \textit{\textbf{Capabilities}}: \textit{Firstly}, this method efficiently updates the DPFL model with new data. \textit{Secondly}, the forgetting factor helps manage the AC power flow model's nonlinearity in non-stationary environments, making training more responsive to recent changes.

\noindent \textit{\textbf{Limitations}}: This method has several common limitations, namely \textbf{L1}, \textbf{L2}, \textbf{L3}, and \textbf{L6}.

\subsubsection{Partial Least Squares and Its Variants}\label{sec:PLS} 
Partial least squares methods are also widespread in DPFL \cite{19, 10, 20, 5, 17}, mainly because they can handle the multicollinearity issue \cite{wold1984collinearity, 19, de1993simpls}. Below are further details and discussions on these algorithms. 

\paragraph{Ordinary Partial Least Squares}~\\

To resolve the singularity caused by the multicollinearity problem, the ordinary partial least squares approach projects $\boldsymbol{X}$ and $\boldsymbol{Y}$ onto some lower dimensional spaces defined by their orthogonal score vectors, thereby removing the correlated components within the original datasets. The projection amounts to decomposing $\boldsymbol{X}$ and $\boldsymbol{Y}$ into
\begin{align}
  \boldsymbol{X} & =  \boldsymbol{T}\boldsymbol{C}^{\top} + \boldsymbol{E} \label{eq:X_PLS} \\
  \boldsymbol{Y} & = \boldsymbol{U}\boldsymbol{R}^{\top} + \boldsymbol{F} \label{eq:Y_PLS}
\end{align}
where $\boldsymbol{T} \in \mathbb{R}^{N_s \times N_p}$ and $\boldsymbol{U} \in \mathbb{R}^{N_s \times N_p}$ consist of $N_p$ score components extracted from $\boldsymbol{X}$ and $\boldsymbol{Y}$;
$\boldsymbol{C} \in \mathbb{R}^{N_x \times N_p}$ and $\boldsymbol{R} \in \mathbb{R}^{N_y \times N_p}$ denote the loading matrices of $\boldsymbol{X}$ and $\boldsymbol{Y}$; $\boldsymbol{E} \in \mathbb{R}^{N_s \times N_x}$ and $\boldsymbol{F} \in \mathbb{R}^{N_s \times N_y}$ represent the matrices of residuals for $\boldsymbol{X}$ and $\boldsymbol{Y}$. The solution of the ordinary partial least squares regression can be explicitly given by \cite{19_26, 19}:
\begin{align}
  \hat{\boldsymbol{\beta}} = \boldsymbol{X}^{\top}\boldsymbol{U}\left(\boldsymbol{T}^{\top}\boldsymbol{X}\boldsymbol{X}^{\top}\boldsymbol{U} \right)^{-1}\boldsymbol{T}^{\top}\boldsymbol{Y} \label{eq:PLS_solu}
\end{align}
{\color{black}
Note that when $N_p = N_x$, i.e., the number of score components is equal to the number of variables in $\boldsymbol{X}$, the ordinary partial least squares regression degrades to the ordinary least squares regression \cite{qin1993partial}. 
}

As for the decomposition algorithms to realize \eqref{eq:X_PLS} and \eqref{eq:Y_PLS}, the most classic one is the nonlinear iterative partial least squares (NIPALS) method proposed in \cite{wold1975path}. To speed up the computation of NIPALS and meanwhile lower its memory requirements \cite{alin2009comparison}, \cite{de1993simpls} presents a statistically inspired modification of the partial least squares (SIMPLS). The SIMPLS algorithm has been employed for DPFL \cite{10, 19, 17}. 



{\color{black}
\noindent \textit{\textbf{Capabilities}}: This method is specifically tailored to address the multicollinearity issue. 

\noindent \textit{\textbf{Limitations}}: \textit{Firstly}, this method necessitates the pre-setting of the hyperparameter $N_p$ (this value can significantly impact the method's performance, particularly when employing SIMPLS for decomposition). This requirement introduces a computationally expensive hyperparameter tuning process. \textit{Secondly}, this method has several typical limitations, including \textbf{L2}, \textbf{L3}, \textbf{L4}, \textbf{L5}, and \textbf{L6}.

}

\paragraph{Recursive Partial Least Squares}\label{sec:RPLS}~\\

To efficiently update the DPFL model with new data, \cite{17} adopts a recursive partial least squares algorithm and applies it to DPFL. Suppose that $\boldsymbol{X}[t]$ and $\boldsymbol{Y}[t] $ have already been decomposed into
\begin{align}
  \boldsymbol{X}[t] & =  \boldsymbol{T}[t]\boldsymbol{C}[t]^{\top} + \boldsymbol{E}[t] \label{eq:X_PLS_k} \\
  \boldsymbol{Y}[t] & = \boldsymbol{U}[t]\boldsymbol{R}[t]^{\top} + \boldsymbol{F}[t] \label{eq:Y_PLS_k}
\end{align}
with
\begin{align}
  \boldsymbol{T}[t] & = \left[\boldsymbol{t}[t]_1 \ \cdots  \ \boldsymbol{t}[t]_{N_p}\right]   \\ 
  \boldsymbol{U}[t] & = [\boldsymbol{u}[t]_1 \ \cdots  \ \boldsymbol{u}[t]_{N_p}]  
\end{align}
When the new measurements at time step $t+1$, i.e., $\boldsymbol{x}_{t+1}$ and $\boldsymbol{y}_{t+1}$, are available, the ordinary partial least squares algorithm has to re-decompose the following datasets 
\begin{align}
  \boldsymbol{X}[t+1] = \left[ \begin{array}{c}
    \boldsymbol{X}[t] \\
    \boldsymbol{x}_{t+1} \\
    \end{array} \right],\ \boldsymbol{Y}[t+1] = \left[ \begin{array}{c}
      \boldsymbol{Y}[t] \\
      \boldsymbol{y}_{t+1} \\
      \end{array} \right]  
\end{align}
in order to generate the latest DPFL model. In contrast, the recursive partial least squares algorithm only needs to decompose
\begin{align}
  \boldsymbol{\widetilde{X}}[t+1] = \left[ \begin{array}{c}
    \boldsymbol{C}[t]^{\top} \\
    \boldsymbol{x}_{t+1} \\
    \end{array} \right],\ \boldsymbol{\widetilde{Y}}[t+1] = \left[ \begin{array}{c}
      \boldsymbol{\varGamma}[t] \boldsymbol{R}[t]^{\top} \\
      \boldsymbol{y}_{t+1} \\
      \end{array} \right]  \label{eq:RPLS_decom}
\end{align}
where 
\begin{align} 
\boldsymbol{\varGamma}[t] = diag(\gamma[t]_{1} \ \cdots \ \gamma[t]_{N_p})
\end{align}
with 
\begin{align} 
\gamma[t]_i = \frac{\boldsymbol{u}[t]_i^{\top}\boldsymbol{t}[t]_i}{\boldsymbol{t}[t]_i^{\top}\boldsymbol{t}[t]_i} 
\end{align}
We denote the decomposition results by 
\begin{align}
  \widetilde{\boldsymbol{X}}[t+1] & =  \widetilde{\boldsymbol{T}}[t+1]\widetilde{\boldsymbol{C}}[t+1]^{\top} + \widetilde{\boldsymbol{E}}[t+1] \label{eq:X_PLS_k_1} \\
  \widetilde{\boldsymbol{Y}}[t+1] & = \widetilde{\boldsymbol{U}}[t+1]\widetilde{\boldsymbol{R}}[t+1]^{\top} + \widetilde{\boldsymbol{F}}[t+1] \label{eq:Y_PLS_k_1}
\end{align}
Substituting $\widetilde{\boldsymbol{X}}[t+1]$ and $\widetilde{\boldsymbol{Y}}[t+1]$ as well as $\widetilde{\boldsymbol{T}}[t+1]$ and $\widetilde{\boldsymbol{U}}[t+1]$ into \eqref{eq:PLS_solu} generates the up-to-date regression coefficients. 

There are several noteworthy points: (i) The recursive partial least squares method is based on the NIPALS algorithm \cite{17_30}; thus, using NIPALS for the decompositions in \eqref{eq:X_PLS_k}, \eqref{eq:Y_PLS_k}, \eqref{eq:X_PLS_k_1}, and \eqref{eq:Y_PLS_k_1} is recommended. (ii) Using NIPALS, the decompositions in \eqref{eq:RPLS_decom} are equivalent to the original decompositions of $\boldsymbol{X}[t+1]$ and $\boldsymbol{Y}[t+1]$ (see \cite{17, 17_30} for details).

\noindent \textit{\textbf{Capabilities}}: \textit{Firstly}, this method efficiently updates the DPFL model with new data. Regardless of $t$, $\widetilde{\boldsymbol{X}}[t+1]$ and $\widetilde{\boldsymbol{Y}}[t+1]$ always have $N_p+1$ rows, leading to a fixed decomposition cost. In contrast, $\boldsymbol{X}[t+1]$ and $\boldsymbol{Y}[t+1]$ have $t+1$ rows, growing linearly with $t$ and reaching $N_s$. Since $N_p \ll N_s$ generally holds, the recursive partial least squares algorithm is much less time-consuming. \textit{Secondly}, this method can handle multicollinearity.

\noindent \textit{\textbf{Limitations}}: This method has the following common limitations: \textbf{L2}, \textbf{L3}, \textbf{L5}, and \textbf{L6}.

\subsubsection{Ridge Regression and Its Variants}
An alternative way to address multicollinearity is to use ridge regression approaches \cite{hoerl1970ridge}, which are also commonly used in DPFL \cite{6,24,4}. These algorithms are further detailed and discussed in the following.

\paragraph{Ordinary Ridge Regression}\label{sec:rr}~\\

The ordinary ridge regression method tunes the diagonal elements of the nearly-singular matrix to enable invertibility by adding a regularization penalty (Tikhonov–Phillips regularization \cite{24}) to the objective function \cite{6}:
\begin{align}
  \min \limits_{\boldsymbol{\beta}} \   \Vert \boldsymbol{Y}- \boldsymbol{X}\boldsymbol{\beta} \Vert_{F}^2 +  \lambda\Vert \boldsymbol{\beta} \Vert_{F}^2   \label{eq:Ridge}
\end{align}
where $\lambda \in \mathbb{R}$ is a tuning parameter. The solution for $\boldsymbol{\beta}$ is
\begin{align}
  \hat{\boldsymbol{\beta}} = \left(\boldsymbol{X}^{\top} \boldsymbol{X} + \lambda\boldsymbol{I} \right)^{-1} \boldsymbol{X}^{\top}\boldsymbol{Y} \label{eq:beta_esti_rigi}
\end{align} 
where $\boldsymbol{I}$ is an identity matrix. The parameter $\lambda$ adjusts the diagonal elements in $\boldsymbol{X}^{\top} \boldsymbol{X}$ to ensure invertibility. In \cite{6}, the intercept in $\boldsymbol{X}$ has been removed. Note that $\lambda$ introduces a bias, so it must be carefully tuned to balance bias and invertibility. See \cite{24, 24_18, 4, 34} for tuning approaches.

\noindent \textit{\textbf{Capabilities}}: \textit{Firstly}, this method is easy to implement and addresses multicollinearity effectively. \textit{Secondly}, it provides a clear and interpretable solution.

\noindent \textit{\textbf{Limitations}}: \textit{Firstly}, tuning the hyperparameter $\lambda$ increases computational load due to repeated training. \textit{Secondly}, this method has several common limitations, including \textbf{L2}, \textbf{L3}, \textbf{L4}, \textbf{L5}, and \textbf{L6}.

\paragraph{Locally Weighted Ridge Regression}~\\

To handle the inherent nonlinearity in AC power flows, \cite{24} proposes a locally weighted ridge regression approach. This method trains local models around target operating points and emphasizes data points closer to the operating point of interest. The objective function becomes
\begin{align}
  \min \limits_{\boldsymbol{\beta}} \    
  \Vert \boldsymbol{W}^{\frac{1}{2}} (\boldsymbol{Y}- \boldsymbol{X}\boldsymbol{\beta}) \Vert_{F}^2 +  \lambda\Vert \boldsymbol{\beta} \Vert_{F}^2  \label{eq:LWRidge}
\end{align}
where $\boldsymbol{W}$ is a diagonal weight matrix. For the power flow model at time step $t$, the $i$-th diagonal element of $\boldsymbol{W}$ is defined as \cite{24, 24_5}
\begin{align} 
  w_i = e^{-\frac{d_i^2}{2\tau^2}}, \ \text{with} \  d_i = \Vert \boldsymbol{x}_{i} - \boldsymbol{x}_{t} \Vert_F, 
\end{align}
where $\tau \in \mathbb{R}$ is tuned by cross-validation \cite{24}. The solution to \eqref{eq:LWRidge} is
\begin{align}
  \hat{\boldsymbol{\beta}} = \left(\boldsymbol{X}^{\top} \boldsymbol{W}\boldsymbol{X} + \lambda\boldsymbol{I} \right)^{-1} \boldsymbol{X}^{\top}\boldsymbol{W}\boldsymbol{Y} \label{eq:beta_esti_rigi_LW}
\end{align} 
In \cite{24, 24_5}, the intercept in $\boldsymbol{X}$ has been removed.

\noindent \textit{\textbf{Capabilities}}: \textit{Firstly}, this method is user-friendly and resolves multicollinearity. \textit{Secondly}, it generates a straightforward, easily understandable solution. \textit{Thirdly}, it tackles the inherent nonlinearity of the AC power flow model.

\noindent \textit{\textbf{Limitations}}: \textit{Firstly}, beyond tuning $\lambda$, this method requires adjusting $\tau$ through cross-validation \cite{24}, increasing computational load. \textit{Secondly}, it has common limitations: \textbf{L2}, \textbf{L3}, \textbf{L4}, and \textbf{L6}.

\paragraph{Clustering-based Ridge Regression}~\\

The clustering-based ridge regression method is another option for handling the nonlinear nature of AC power flows \cite{4}. This method produces a piecewise linear model that better fits the nonlinear hypersurface. Specifically, the K-plane clustering method proposed in \cite{4_35, 4_26} is integrated into ridge regression \cite{4}. This approach fits the AC power flow model piecewise according to \textbf{Model 2}. Unlike the clustering-based least squares method, ridge regression is embedded into each iteration of the clustering procedure. The steps are summarized below:
\begin{enumerate}
  \item[(i)] Initialize 
  $[\boldsymbol{S}(k)^n, \boldsymbol{\mu}(k)^n,\hat{\boldsymbol{\beta}}(k)^n ]$ for $n=0$ and $k=1\ \cdots \ K$, where $n$ indicates the iteration number, $\boldsymbol{S}(k)^n$ consists of $\boldsymbol{x}_i$ belonging to cluster $k$, $\boldsymbol{\mu}(k)^n$ denotes the centroid of cluster $k$, and $\hat{\boldsymbol{\beta}}(k)^n$ represents the regression coefficient for cluster $k$.
  \item[(ii)] Update $\hat{\boldsymbol{\beta}}(k)^n$  by 
  $$\hat{\boldsymbol{\beta}}(k)^{n+1} = \left[\boldsymbol{X}(k)^{\top} \boldsymbol{X}(k) + \lambda\boldsymbol{I} \right]^{-1} \boldsymbol{X}(k)^{\top} \boldsymbol{Y}(k), \ \forall k $$ where $\boldsymbol{X}(k)$ is composed of $\boldsymbol{x}_i \in \boldsymbol{S}(k)^n $ while $\boldsymbol{Y}(k)$ collects $\boldsymbol{y}_i$ that corresponds to $\boldsymbol{x}_i \in \boldsymbol{S}(k)^n$. 
  \item[(iii)] Update $\boldsymbol{\mu}(k)^n$  according to
  $$\boldsymbol{\mu}(k)^{n+1} =  \sum\nolimits_{\boldsymbol{x}_i \in \boldsymbol{S}(k)^n} \boldsymbol{x}_i/|\boldsymbol{S}(k)^n|, \ \forall k   $$
  where $|\cdot|$ denotes the cardinality function. 
  \item[(iv)] Re-allocate $\boldsymbol{x}_i \ (\forall i)$ to cluster $k$, where
  $$ 
  k = arg \min \limits_{j} \   \Vert  \boldsymbol{y}_i^{\top} - \boldsymbol{x}_i^{\top}\hat{\boldsymbol{\beta}}(j)^{n+1}\Vert_F^2 + \eta\Vert \boldsymbol{x}_i - \boldsymbol{\mu}(j)^{n+1}\Vert_F^2  
  $$
Parameter $\eta \in \mathbb{R}$ is specified using cross-validation \cite{4}. Update $\boldsymbol{S}(k)^n$ to $\boldsymbol{S}(k)^{n+1}$ for $\forall k$ accordingly. 
  \item[(v)] Stop if $\boldsymbol{S}(k)^{n+1} =  \boldsymbol{S}(k)^{n} ~(\forall k) $ and output $$\boldsymbol{\mu}(k) = \boldsymbol{\mu}(k)^{n+1}, \ \hat{\boldsymbol{\beta}}(k) = \hat{\boldsymbol{\beta}}(k)^{n+1}, \ \forall k$$
  Otherwise, set $n = n+1$ and go to step (ii).
\end{enumerate}

To apply the obtained piecewise linear power flow model, one should first locate which cluster a given input $\boldsymbol{x}$ belongs to by 
$$ k = arg \min \limits_{j} \Vert \boldsymbol{x} - \boldsymbol{\mu}(j)\Vert_F^2 $$
Then, $\boldsymbol{x}$ can be substituted into $\hat{\boldsymbol{y}} = \hat{\boldsymbol{\beta}}(k)^{\top}\boldsymbol{x}$ to predict the value of $\boldsymbol{y}$.

\noindent \textit{\textbf{Capabilities}}: \textit{Firstly}, this method addresses the AC power flow model's nonlinearity by incorporating clustering, potentially enhancing DPFL model accuracy. It uses the K-plane clustering method, aligning better with the linearization objective than K-means or Gaussian mixture models. \textit{Secondly}, it can handle multicollinearity issues.

\noindent \textit{\textbf{Limitations}}: \textit{Firstly}, incorporating $\Vert \boldsymbol{x}_i - \boldsymbol{\mu}(j)^{n+1}\Vert_F^2$ in the cost function adds bias to $\boldsymbol{\beta}$ since $\hat{\boldsymbol{\beta}}(j)$ depends on $\boldsymbol{\mu}(j)$. \textit{Secondly}, tuning the hyperparameter $\lambda$, the number of clusters, and $\eta$ through cross-validation increases computational effort. Tuning $\eta$ is challenging due to the different magnitudes of $\Vert \boldsymbol{y}_i^{\top} - \boldsymbol{x}_i^{\top}\hat{\boldsymbol{\beta}}(j)^{n+1}\Vert_F^2$ and $\Vert \boldsymbol{x}_i - \boldsymbol{\mu}(j)^{n+1}\Vert_F^2$, sometimes making $\eta$ redundant. \textit{Thirdly}, allocating inputs to clusters based solely on proximity to centroids, without considering their relationship to hyperplanes, can result in suboptimal outcomes. \textit{Fourthly}, using many clusters can lead to underdetermined linear systems when observations in a cluster are fewer than $N_y$, yielding invalid results. \textit{Fifthly}, as a piecewise linear model, integrating the output into decision-making frameworks like optimal power flow problems requires extra effort, potentially introducing additional integer variables and increasing complexity. \textit{Lastly}, this method also has common limitations \textbf{L2}, \textbf{L3}, \textbf{L4}, and \textbf{L6}.

\subsubsection{Support Vector Regression and Its Variants}
The support vector regression method and its variants have also been used in \cite{14, 20, 27, 28} for DPFL studies. These methods are further discussed below. 

\paragraph{Ordinary Support Vector Regression}~\\

The ordinary support vector regression method uses
\begin{align} 
\max\left\{ 0, \Vert y_{ij} - \boldsymbol{x}_i^{\top}\boldsymbol{\beta}_j  \Vert_1 - \epsilon  \right\}
\end{align}
in its objective function. This term is called the $\epsilon$-insensitive error function, meaning that only absolute errors greater than $\epsilon$ are minimized, while smaller errors and corresponding data points are neglected. Compared to other regression methods, the $\epsilon$-insensitive error function offers higher tolerance for minor data outliers \cite{14,27,28}. 

However, the $\epsilon$-insensitive error function is not differentiable everywhere. Slack variables are thus introduced for relaxation, turning the regression model into:
\begin{equation}
  \begin{aligned}
  \min_{\boldsymbol{\beta}, \  \xi_{ij},\  \xi_{ij}^{\star}} \quad  & \frac{1}{2}\Vert \boldsymbol{\beta} \Vert_{F}^2 + \omega\sum\nolimits_{i=1}^{N_s}\sum\nolimits_{j=1}^{N_y} \left( \xi_{ij} +  \xi_{ij}^{\star}\right) \\
  \textrm{s.t.} \quad  & y_{ij}-\boldsymbol{x}_i^{\top}\boldsymbol{\beta}_j \leq \epsilon + \xi_{ij}, \ \forall i, j \\
  & \boldsymbol{x}_i^{\top}\boldsymbol{\beta}_j - y_{ij} \leq \epsilon + \xi_{ij}^{\star}, \ \ \forall i, j \\
  & \xi_{ij}, \xi_{ij}^{\star} \geq 0, \ \ \forall i, j \\
  \end{aligned} \label{eq:SVR}
\end{equation}
where hyperparameters $\omega$ and $\epsilon$ are determined via cross-validation \cite{14, 34}. This model is often solved through its dual problem using the sequential minimal optimization algorithm; see \cite{14_29} for details.

Data multicollinearity is generally not an issue for support vector regression, partly because the inverse calculation of $\boldsymbol{X}^{\top}\boldsymbol{X}$ is not required and partly because this method uses Tikhonov–Phillips regularization, reducing the singularity caused by multicollinearity. While the regularization term coefficient is fixed at 0.5, an additional tunable regularization term can further handle multicollinearity and bad data \cite{34}, i.e.,
$$
\min_{\boldsymbol{\beta}, \  \xi_{ij},\  \xi_{ij}^{\star}}  \frac{1}{2}\Vert \boldsymbol{\beta} \Vert_{F}^2 + \omega\sum\nolimits_{i=1}^{N_s}\sum\nolimits_{j=1}^{N_y} \left( \xi_{ij} +  \xi_{ij}^{\star}\right) + \lambda\Vert \boldsymbol{\beta} \Vert_{F}^2
$$

{\color{black}
\noindent \textit{\textbf{Capabilities}}: \textit{Firstly}, this method is capable of addressing multicollinearity issues. \textit{Secondly}, it is less sensitive to minor data outliers, which allows for higher generalizability. 

\noindent \textit{\textbf{Limitations}}: \textit{Firstly}, while the support vector regression model is inherently convex, solving it, especially when applied to large-scale power systems, remains a challenge due to the extensive number of decision variables in $\boldsymbol{\beta}$. \textit{Secondly}, the method requires the tuning of hyperparameters $\omega$ and $\epsilon$ through cross-validation, which increases the computational load. Note that $\epsilon$ significantly influences the method's performance. Yet, identifying an appropriate range for tuning $\epsilon$ is not straightforward, as this range is highly dependent on the magnitudes of the electrical measurements, whether and how the measurements are normalized, etc. \textit{Lastly}, this method is also subject to common limitations, including \textbf{L3}, \textbf{L4}, and \textbf{L5}.
}

\paragraph{Support Vector Regression with Kernels}\label{sec:SVR_k}~\\

As mentioned earlier, projecting $\boldsymbol{x}$ and $\boldsymbol{y}$ to other spaces may handle the inherent nonlinearity of the AC power flow model. The Kernel-based support vector regression method has thus been used in DPFL \cite{27,28}. This approach uses \textbf{Model 3} as the base linear power flow model. Specifically, each realization of $\boldsymbol{x}$, i.e., $\boldsymbol{x}_i$, is projected to an $N_{\phi}$-dimensional space via the mapping function $\phi(\boldsymbol{x}_i)$. The original regression model in \eqref{eq:SVR} then becomes:
\begin{equation}
  \begin{aligned}
  \min_{\boldsymbol{\beta}_{\phi}, \  \xi_{ij},\  \xi_{ij}^{\star}} \quad  & \frac{1}{2}\Vert  \boldsymbol{\beta}_{\phi} \Vert_{F}^2 + \omega\sum\nolimits_{i=1}^{N_s}\sum\nolimits_{j=1}^{N_y} \left( \xi_{ij} +  \xi_{ij}^{\star}\right) \\
  \textrm{s.t.} \quad  & y_{ij}-\phi(\boldsymbol{x}_i)^{\top}\boldsymbol{\beta}_{\phi j} \leq \epsilon + \xi_{ij}, \ \forall i, j \\
  & \phi(\boldsymbol{x}_i)^{\top}\boldsymbol{\beta}_{\phi j} - y_{ij} \leq \epsilon + \xi_{ij}^{\star}, \ \ \forall i, j \\
  & \xi_{ij}, \xi_{ij}^{\star} \geq 0, \ \ \forall i, j \\
  \end{aligned} \label{eq:SVR_k}
\end{equation}
where $\boldsymbol{\beta}_{\phi j}$ refers to the $j$-th column of $\boldsymbol{\beta}_{\phi}$. 

Note that calculating $\left\langle \phi(\boldsymbol{x}_i), \phi(\boldsymbol{x}_j) \right\rangle$ for $\forall i, j$ is necessary when solving \eqref{eq:SVR_k}. These inner products are challenging to compute because the dimension of $\phi(\boldsymbol{x}_i)$, $N_{\phi}$, is usually high. However, if $\phi(\cdot)$ is chosen from the reproducing Hilbert kernel space (RHKS), these inner products can be easily calculated due to the property \cite{27, 28}:
\begin{align}
  \left\langle \phi(\boldsymbol{x}_i), \phi(\boldsymbol{x}_j) \right\rangle = h\left( \left\langle \boldsymbol{x}_i, \boldsymbol{x}_j \right\rangle \right)
\end{align}
where $h(\cdot)$ is a scalar function. This property allows high-dimensional inner products to be computed using low-dimensional calculations, i.e., $h\left( \left\langle \boldsymbol{x}_i, \boldsymbol{x}_j \right\rangle \right)$. Define $\kappa(\boldsymbol{x}_i, \boldsymbol{x}_j) \coloneqq h\left( \left\langle \boldsymbol{x}_i, \boldsymbol{x}_j \right\rangle \right)$, where $\kappa(\boldsymbol{x}_i, \boldsymbol{x}_j)$ is the kernel function.

Using a kernel from RHKS, the dual problem of \eqref{eq:SVR_k} can be solved using the sequential minimal optimization algorithm \cite{flake2002efficient}. Correspondingly, \textbf{Model 3} in \eqref{eq:linear_map} becomes
\begin{align}
  \hat{\boldsymbol{y}} = \hat{\boldsymbol{\beta}}_{\phi}^{\top}\phi(\boldsymbol{x}) = \sum\nolimits_{i=1}^{N_s}\alpha_i^{\star} \boldsymbol{y}_i \kappa(\boldsymbol{x}_i, \boldsymbol{x}) \label{eq:linear_map_details} 
\end{align}
where $\alpha_i^{\star} \ (\forall i)$ are the optimal Lagrange multipliers of \eqref{eq:SVR_k}; see \cite{27_15} for more details. The application of \eqref{eq:linear_map_details} is straightforward: for any given input $\boldsymbol{x}$, compute $\kappa(\boldsymbol{x}_i, \boldsymbol{x})$ for $\forall i$, then use these kernel values and historical measurements $\boldsymbol{y}_i~(\forall i)$ in \eqref{eq:linear_map_details} to estimate $\boldsymbol{y}$.

\noindent \textit{\textbf{Capabilities}}: \textit{Firstly}, this method addresses multicollinearity. \textit{Secondly}, it is less sensitive to minor data outliers, allowing for higher generalizability. \textit{Thirdly}, it handles the inherent nonlinearity of the power flow model.

\noindent \textit{\textbf{Limitations}}: \textit{Firstly}, kernel selection is restricted to RHKS. Common kernels (e.g., linear, polynomial, Gaussian, sigmoid) may not always create a space where $\boldsymbol{y}$ and projected $\boldsymbol{x}$ are highly linear. \textit{Secondly}, the linear power flow model from \eqref{eq:linear_map_details} is not a linear function of $\boldsymbol{x}$ but of $\kappa(\boldsymbol{x}_i, \boldsymbol{x})$, limiting its application. \textit{Thirdly}, solving the optimization problem involves many decision variables (i.e., $\boldsymbol{\beta}_{\phi}$), especially in large-scale power systems, leading to high computational demands. \textit{Fourthly}, tuning hyperparameters $\omega$ and $\epsilon$ via cross-validation further increases computational effort. \textit{Lastly}, this method has common limitations \textbf{L3} and \textbf{L4}.

\subsection{Tailored Methods}
The tailored DPFL training algorithms proposed in \cite{3,9,11,33} are discussed below. Each method is referred to by the type of programming model it uses, as no specific names exist for these algorithms.

\subsubsection{Linearly Constrained Programming}
Several researchers advocate for integrating physical knowledge into the data-driven training process to enhance DPFL model performance. One approach is adding physical constraints to DPFL programming models \cite{9, 11}. The physical constraints used include:
\begin{itemize}
  \item Bound constraints for $\boldsymbol{\beta}$.
  \item Structure constraints for $\boldsymbol{\beta}$.
  \item Coupling constraints for elements in $\boldsymbol{\beta}$.
\end{itemize}
These constraints are discussed below. 

\paragraph{Bound Constraints}~\\

Every power system has operational boundary conditions. These conditions can be used to estimate the upper and lower bounds of $\boldsymbol{\beta}$ using physics-driven linearization methods, such as the first-order Taylor series expansion around the boundary operational points. The resulting coefficients can be considered as the limits of $\boldsymbol{\beta}$ \cite{9, 25, shao2023physical}, i.e.,
\begin{align} 
\underline{ \boldsymbol{\beta}} \leq \boldsymbol{\beta} \leq \overline{ \boldsymbol{\beta}}
\end{align}
where $\overline{ \boldsymbol{\beta}}$ is the upper limit and $\underline{ \boldsymbol{\beta}}$ is the lower limit. These constraints can be incorporated into the data-driven training process.

\paragraph{Structure Constraints}~\\

Under some assumptions, a similarity between $\boldsymbol{\beta}$ and the Jacobian matrix from the first-order Taylor series expansion has been observed in \cite{19} and elaborated in \cite{11}. This similarity enforces $\boldsymbol{\beta}$ to follow a specific structure. Specifically, assuming $\sin \theta_{ij} \ll \cos \theta_{ij}$ in power systems, the terms with $\sin \theta_{ij}$ in the Jacobian matrix can be neglected compared to those with $\cos \theta_{ij}$ \cite{11}. This simplification results in a symmetrical Jacobian matrix, and thus, $\boldsymbol{\beta}$ is assumed to follow this structure \cite{11}, i.e.,
\begin{align}
  \boldsymbol{\beta} = \left[ 
    \begin{array}{cc}
    \boldsymbol{\beta}_{B} & -\boldsymbol{\beta}_{G}  \\
    \boldsymbol{\beta}_{G} & \ \ \boldsymbol{\beta}_{B} \\
    \end{array} 
    \right] +
    \left[ 
    \begin{array}{cc}
      -\boldsymbol{\beta}_{Q} & \boldsymbol{\beta}_{P}  \\
      \ \ \boldsymbol{\beta}_{P} & \boldsymbol{\beta}_{Q} \\
    \end{array} 
    \right] \label{eq:structure_1}
\end{align}
where
\begin{align}
  \boldsymbol{\beta}_{B} = \boldsymbol{\beta}_{B}^{\top},\  \boldsymbol{\beta}_{G} = \boldsymbol{\beta}_{G}^{\top}, \label{eq:structure_2}
\end{align}
with $\boldsymbol{\beta}_{Q}$ and  $\boldsymbol{\beta}_{P}$ being two diagonal matrices, $\boldsymbol{\beta}_{B} = \boldsymbol{\beta}_{B}^{\top}$, and  $\boldsymbol{\beta}_{G} = \boldsymbol{\beta}_{G}^{\top}$. The above structure implies symmetry and diagonality, which can be enforced by several linear constraints on $\boldsymbol{\beta}$. These constraints are also known as the Jacobian-matrix-guided constraints \cite{11}. 



\paragraph{Coupling Constraints}~\\

In some cases, the elements in $\boldsymbol{\beta}$ are highly related due to physical dependencies in power systems. For example, when training a DPFL model between the active power flow \( P_{nm} \) and the voltages/angles of buses \( n \) and \( m \), the angles in the AC line flow model appear as \( \theta_{nm}=\theta_n - \theta_m \). Thus, the coefficients for \( \theta_n \) and \( \theta_m \) in the DPFL model should be exactly opposite, i.e., \( \beta_{nj} + \beta_{mj} = 0 \). This constraint can be slightly relaxed for practical use \cite{9}:
\begin{align}
  -\delta^{LIN} \leq \beta_{nj} + \beta_{mj} \leq \delta^{LIN}  \label{eq:lin1}
\end{align}
where \( \delta^{LIN} \in \mathbb{R} \) is a preset, non-negative small value.

{\color{black}
\noindent \textit{\textbf{Capabilities}}: \textit{Firstly}, these linear constraints keep the quadratic regression models convex. The optimization problems can be solved by commercial solvers like MOSEK or GUROBI, or accelerated algorithms like the Adam-DSGD method \cite{liu2023data}. \textit{Secondly}, integrating physical knowledge into the DPFL training process can enhance model performance, especially with accurate physical parameters. \textit{Thirdly}, linearly-constrained DPFL methods are generally less sensitive to multicollinearity.

\noindent \textit{\textbf{Limitations}}: \textit{Firstly}, adding physical knowledge may lead to the loss of other vital details. E.g., bound and structure constraints rely on the Jacobian matrix from the classic first-order Taylor approximation method, limiting predictors to known active/reactive power injections and responses to unknown voltages/angles. This excludes known voltages from PV buses, losing significant information. \textit{Secondly}, in power systems, normalizing variables to a common range can mitigate training issues (e.g., numerical instability). Yet, integrating physical knowledge may be problematic for normalized datasets, especially with variance-scaling, because such independent normalization can invalidate physical relationships in power flow equations. \textit{Thirdly}, the benefit of integrating physical knowledge depends on its accuracy. The Jacobian matrix, for example, used as the physical knowledge base for $\boldsymbol{\beta}$, is not the ground-truth linear power flow model and may not enhance the training accuracy. \textit{Fourthly}, the constraints limit the choice of responses, making the gained DPFL model restricted to certain variables (e.g., only the unknown voltages/angles). Developing a model that can predict all unknown variables (e.g., branch flows) requires additional linearization methods. \textit{Lastly}, linearly-constrained DPFL methods have common drawbacks, including \textbf{L2}, \textbf{L4}, \textbf{L5}, and \textbf{L6}.

}

\subsubsection{Chance-constrained Programming} 
The hyperparameters in support vector regression, e.g., $\omega$ in \eqref{eq:SVR}, are challenging to tune \cite{33}. To avoid tuning these hyperparameters, a chance-constrained programming method has been proposed for DPFL training in \cite{33}. This method removes the tolerance of residuals (the second term of the objective in \eqref{eq:SVR} and the corresponding constraints). Instead, single-chance constraints are added to restrict residuals.  The resulting model is given by
\begin{gather}
  \min_{\boldsymbol{\beta}} \   \frac{1}{2}\Vert \boldsymbol{\beta} \Vert_{F}^2   \label{eq:CCP} \\
  \textrm{s.t.} \   \mathbb{P}\left\{ \Vert \boldsymbol{\mathcal{Y}}_{j} - \boldsymbol{\mathcal{X}}^{\top}\boldsymbol{\beta}_j  \Vert_1 \leq \epsilon \right\} \geq \zeta^{CCP}_{j}, \  \forall j  \notag 
\end{gather}
where $\boldsymbol{\mathcal{X}} \in  \mathbb{R}^{N_x \times 1}$ rather than $\boldsymbol{x}$ is adopted, aiming to highlight that $\boldsymbol{x}$ is now a random variable. In addition, $\boldsymbol{\mathcal{Y}}_{j} \in  \mathbb{R}$ refers to the $j$-th dependent variable in $\boldsymbol{y}$, which should also be considered as a random variable, owing to the randomness of $\boldsymbol{\mathcal{X}}$ and the physical relationship between $\boldsymbol{\mathcal{Y}}_{j}$ and $\boldsymbol{\mathcal{X}}$. Further, the constant $\zeta^{CCP}_{ij} \in \mathbb{R}$ is a predefined percentage threshold. Additionally, one can add a constraint $\boldsymbol{\beta} \in \boldsymbol{\mathscr{P}} $ into the above chance-constrained model to restrict the bounds of $\boldsymbol{\beta}$, where set $\boldsymbol{\mathscr{P}}$ denotes a physical range of $\boldsymbol{\beta}$ \cite{33}, e.g., the bound and coupling constraints discussed earlier. 


To solve \eqref{eq:CCP}, the big-M method is used in \cite{33} to convert the model into a mixed-integer linear programming problem based on scenarios from $\boldsymbol{\mathcal{Y}}_{j}$ and $\boldsymbol{\mathcal{X}}$. These scenarios are taken from the datasets $\boldsymbol{Y}$ and $\boldsymbol{X}$. The converted problem can be efficiently solved using commercial solvers like GUROBI or CPLEX.

\noindent \textit{\textbf{Capabilities}}: \textit{Firstly}, this method eliminates a hyperparameter, reducing the computational effort for tuning. \textit{Secondly}, it is less sensitive to minor data outliers, leading to better generalizability. \textit{Thirdly}, $\epsilon$ is more straightforward to interpret as the upper bound of the linearization error within the confidence interval across all training samples, making it easier to tune. In standard support vector regression, $\epsilon$ determines which samples fall outside the $\epsilon$-insensitive zone, making it less intuitive to adjust. \textit{Lastly}, this method is less vulnerable to multicollinearity.

\noindent \textit{\textbf{Limitations}}: \textit{Firstly}, chance constraints allow violations with a probability of $1-\zeta^{CCP}_{j}$. Hence, significant linearization errors can still occur without a guarantee of the worst-case performance. \textit{Secondly}, if $\zeta^{CCP}_{j}$ is set close to 1, the method may become infeasible, unless $\epsilon$ is impractically large. \textit{Thirdly}, the big-M method introduces many additional constraints, increasing the problem's scale. \textit{Lastly}, this method also shares common limitations, specifically \textbf{L3}, \textbf{L4}, and \textbf{L5}.

\subsubsection{Distributionally Robust Chance-constrained Programming} 
Most DPFL training algorithms generate linear power flow models by minimizing residuals or using chance constraints, which can result in notable errors in extreme cases. To address this, \cite{3} introduces distributionally robust chance constraints into data-driven training to explicitly limit worst-case errors of the DPFL model. The is given as follows \cite{3}:
\begin{gather}
  \min_{\boldsymbol{\beta}} \   r\left( \boldsymbol{y}^{\star},\boldsymbol{x}^{\star}, \boldsymbol{\beta} \right)  \label{eq:DRCC} \\
  \textrm{s.t.} \   \inf_{\mathbb{P}(\boldsymbol{\mathcal{X}}) \in \boldsymbol{\mathcal{D}}}  \mathbb{P}\left\{ \widetilde{r}_{j} \left( \boldsymbol{\mathcal{Y}}_{j}, \boldsymbol{\mathcal{X}}, \boldsymbol{\beta}_j  \right) \leq \epsilon_j  \right\} \geq \zeta^{DRC}_{j}, \forall j  \notag
\end{gather}
where $r(\cdot)$ is a general description of a residual. Data points $\boldsymbol{y}^{\star} \in  \mathbb{R}^{N_y \times 1} $ and $\boldsymbol{x}^{\star} \in  \mathbb{R}^{N_x \times 1}$ are two given realizations of $\boldsymbol{y}$ and $\boldsymbol{x}$, respectively. This pair of realizations represents a typical operating point \cite{3}. Correspondingly, the physical meaning of the objective function is to minimize the residual of the target linear model around a chosen typical operating point \cite{3}. In addition, $\epsilon_j \in \mathbb{R} $ and $\zeta^{DRC}_{j} \in \mathbb{R}$ are two preset thresholds. Random variables $\boldsymbol{\mathcal{Y}}_{j}$ and $\boldsymbol{\mathcal{X}}$ have the same meaning as in the chance-constrained programming discussed previously. Since it is often difficult to accurately assume a prior probability distribution for random variables, an ambiguity set $\boldsymbol{\mathcal{D}}$ (either moment-based or divergence-based) is utilized to describe multiple possible selections of $\mathbb{P}(\boldsymbol{\mathcal{X}})$, i.e., the probability distribution of $\boldsymbol{\mathcal{X}}$ \cite{3_10}, yielding the distributionally robust chance constraint demonstrated in \eqref{eq:DRCC}. In this constraint, the formulation of $\widetilde{r}_{j}(\cdot)$ is a linear equivalent or approximation of $r(\cdot)$ \cite{3}, e.g., 
\begin{align} 
\widetilde{r}_{j} \left( \boldsymbol{\mathcal{Y}}_{j}, \boldsymbol{\mathcal{X}}, \boldsymbol{\beta}_j  \right) = \Vert    \boldsymbol{\mathcal{Y}}_{j}-\boldsymbol{\mathcal{X}}^{\top}\boldsymbol{\beta}_j \Vert_1, \ \forall j
\end{align}
if 
\begin{align} 
r\left( \boldsymbol{y}^{\star},\boldsymbol{x}^{\star}, \boldsymbol{\beta} \right) = \Vert \boldsymbol{y}^{\star \top} -\boldsymbol{x}^{\star \top}\boldsymbol{\beta} \Vert_F^2
\end{align}

{\color{black}

To solve \eqref{eq:DRCC}, the key is to reformulate the distributionally robust chance constraints. If the ambiguity set $\boldsymbol{\mathcal{D}}_j$ is moment-based, the constraints can be transformed into semi-definite constraints using the conic dual transformation \cite{3_9, jiang2016data}, or into second-order conic constraints \cite{shao2023physical}, which are tractable. Solvers like MOSEK, SeDuMi, and SDPT3 can solve the transformed model. If $\boldsymbol{\mathcal{D}}_j$ is $\phi$-divergence-based, the constraints can be mapped to single-chance constraints using the method in \cite{3_10}, and further transformed into deterministic ones using the big-M approach. Solvers like GUROBI can then solve the resulting model.

\noindent \textit{\textbf{Capabilities}}: \textit{Firstly}, this method explicitly restricts the worst-case errors of the DPFL model. \textit{Secondly}, it is less vulnerable to multicollinearity. \textit{Thirdly}, it provides a universal framework for tuning DPFL model performance using distributionally robust chance constraints.

\noindent \textit{\textbf{Limitations}}: \textit{Firstly}, this method is designed to train a linear power flow model around the operating point $(\boldsymbol{y}^{\star}, \boldsymbol{x}^{\star})$. Its accuracy declines as the state deviates from this point \cite{3}. \textit{Secondly}, the uncertainty in the distribution of $\boldsymbol{\mathcal{Y}}_{j}$ is not considered. The model \eqref{eq:DRCC} only accounts for the uncertainty of $\mathbb{P}(\boldsymbol{\mathcal{X}})$, which may be insufficient. \textit{Thirdly}, overlooking the uncertainty in $\mathbb{P}(\boldsymbol{\mathcal{Y}}_j)$ is questionable, as $\boldsymbol{\mathcal{Y}}_j$ and $\boldsymbol{\mathcal{X}}$ should follow a joint probability distribution $\mathbb{P}(\boldsymbol{\mathcal{X}}, \boldsymbol{\mathcal{Y}}_j)$. The model should consider this joint uncertainty. \textit{Fourthly}, there is no computational advantage. If the ambiguity set $\boldsymbol{\mathcal{D}}_j$ is moment-based and transformed into semi-definite constraints, solving the problem is time-consuming. Substituting with second-order conic constraints may speed up the process but increase the optimality gap. Using the big-M approach can improve efficiency but still add numerous constraints, making the solution time longer than other regression-based DPFL algorithms. \textit{Lastly}, this method shares common limitations, including \textbf{L3}, \textbf{L4}, and \textbf{L5}.

}

\section{Supportive Techniques}\label{sec:tech}
\begin{figure*}[]
	\centering 
	\includegraphics[width=7.3in]{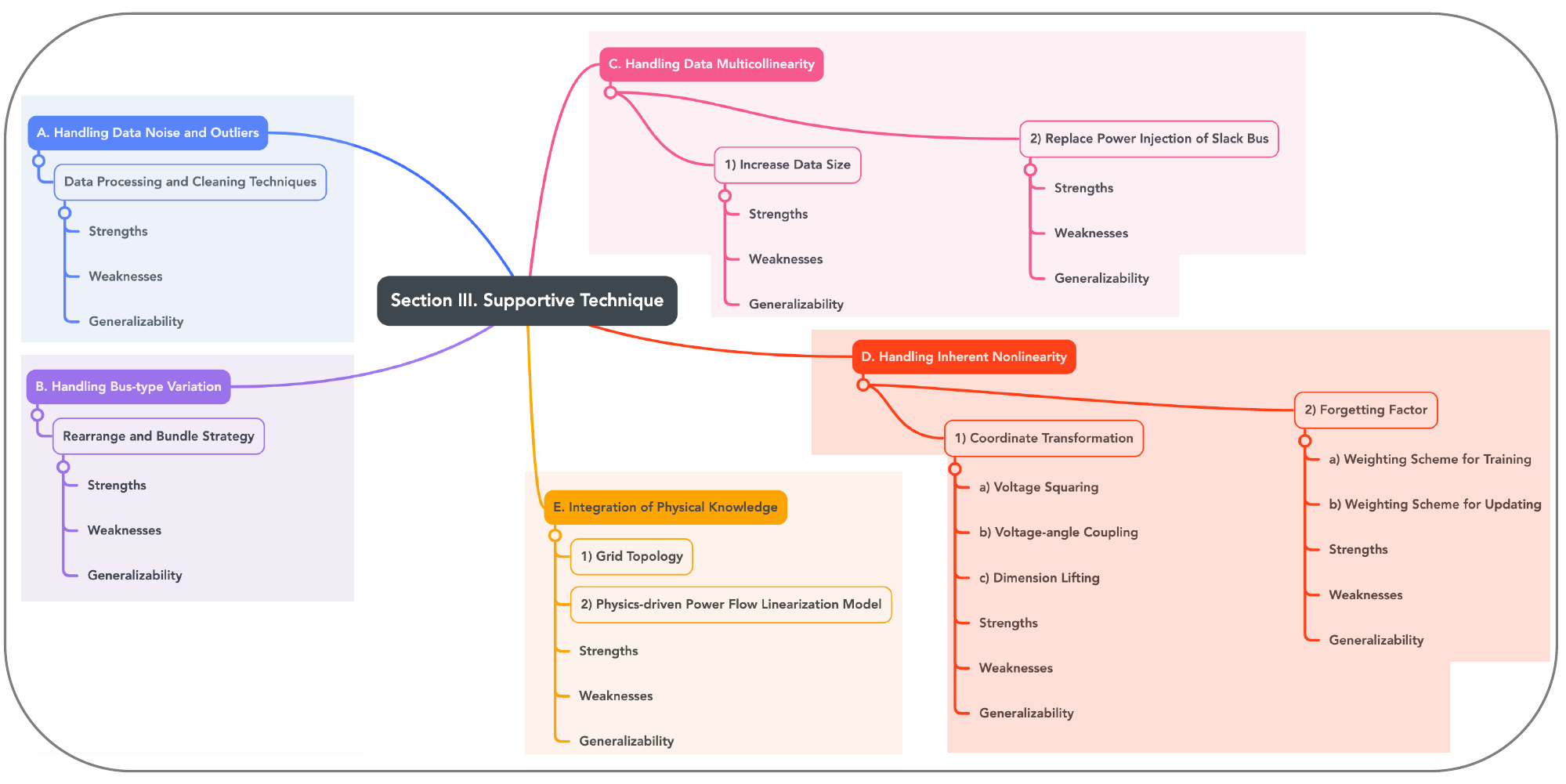} 
	\vspace{-0.4cm}
	\caption{Graphical mirror of the outline of Section III, also explaining the classifications (based on the target) and interconnections of existing DPFL supportive techniques. }
	\label{Fig:Supporttech} 
\end{figure*}

{\color{black} 
To improve the training performance, a variety of DPFL supportive techniques have been employed. These supportive methods, as well as their capabilities, limitations, and generalizability, are discussed and summarized in this section. Fig. \ref{Fig:Supporttech} also provides a graphical overview of the different methods discussed in this section.

}

\subsection{Handling Data Noise and Outliers}\label{sec:noise_outliers}

A variety of data processing and cleaning techniques exist to address issues related to data noise and outliers. For example, the well-known Kalman filter is effective in mitigating noise and handling outliers, as noted in \cite{liu2023data} for the DPFL study.

\noindent \textit{\textbf{Capabilities}}: \textit{Firstly}, numerous techniques can effectively mitigate data noise and outliers, and many of these techniques are accessible through software and open-source code, facilitating ease of use.

\noindent \textit{\textbf{Limitations}}: \textit{Firstly}, the efficacy of these methods depends on deep data comprehension. For example, the Kalman filter requires knowing the transition matrix, the process noise covariance matrix, etc., which are not directly observable in power systems, as endogenous data fluctuations across multiple time scales complicate isolating noise from original measurements.

\noindent \textit{\textbf{Generalizability}}: Data processing and cleaning techniques generally apply to any previously discussed DPFL training algorithms.

\subsection{Handling Bus-type Variation}\label{sec:bus_type_varia}
One challenge facing DPFL models is that they may become invalid when there are variations in bus types. This is because bus-type changes can alter the assignment of known/unknown variables; the new assignment of variables may not match the assignment implicitly used in the learned DPFL model. To deal with the issue of bus-type variations, \cite{19} proposes a bundle strategy for known (a.k.a., independent) and unknown (a.k.a., dependent) variables. Specifically, let us first separate $\boldsymbol{X}$ into $\boldsymbol{X}_1 \in \mathbb{R}^{N_s \times N_{x_1}} $ and $\boldsymbol{X}_2 \in \mathbb{R}^{N_s \times N_{x_2}}$, where $N_x=N_{x_1}+N_{x_2}$. Dataset $\boldsymbol{X}_1$ collects the measurements of the active/reactive power injections at PQ buses and the active power injections at PV buses. Dataset $\boldsymbol{X}_2$ collects the observations of the voltages of the slack and PV buses. Similarly, $\boldsymbol{Y}$ splits into $\boldsymbol{Y}_1 \in \mathbb{R}^{N_s \times N_{y_1}} $ and $\boldsymbol{Y}_2 \in \mathbb{R}^{N_s \times N_{y_2}}$, where $N_y=N_{y_1}+N_{y_2}$. Dataset $\boldsymbol{Y}_1$ contains the realizations of the angles of PQ and PV buses, the voltages of PQ buses, and the active power injection of the slack bus. Dataset $\boldsymbol{Y}_2$ consists of the measurements for the reactive power injections at the slack and PV buses. Note that $N_{x_2} = N_{y_2}$ holds. 

In the strategy developed in \cite{19}, $\boldsymbol{Y}_1$ and $\boldsymbol{X}_2$ are bundled together, and $\boldsymbol{X}_1$ and $\boldsymbol{Y}_2$ are bundled together. Subsequently, this strategy intends to estimate the following relationship:
\begin{align}
  \left[ 
    \begin{array}{cc}
    \boldsymbol{Y}_{1} & \boldsymbol{X}_{2}  \\
    \end{array} 
    \right]  = 
    \left[ 
    \begin{array}{cc}
      \boldsymbol{X}_{1} & \boldsymbol{Y}_{2}  \\
    \end{array} 
    \right] \left[ 
      \begin{array}{cc}
      \boldsymbol{\beta}_{11} & \boldsymbol{\beta}_{12}  \\
      \boldsymbol{\beta}_{21} &  \boldsymbol{\beta}_{22} \\
      \end{array} 
      \right]  \label{eq:rearrange}
\end{align}
using any DPFL training algorithm. In the end, for any given input $\boldsymbol{x} = [\boldsymbol{x}_1^{\top}\ \  \boldsymbol{x}_2^{\top}]^{\top}$, the estimated value of $\boldsymbol{y} = [\boldsymbol{y}_1^{\top}\ \  \boldsymbol{y}_2^{\top}]^{\top}$ can be obtained via
\begin{align}
  \hat{\boldsymbol{y}}_2^{\top} & = \left(\boldsymbol{x}_2^{\top} - \boldsymbol{x}_1^{\top}\hat{\boldsymbol{\beta}}_{12} \right)\hat{\boldsymbol{\beta}}_{22}^{-1} \label{eq:beta_22} \\ 
  \hat{\boldsymbol{y}}_1^{\top} & = \boldsymbol{x}_1^{\top}\hat{\boldsymbol{\beta}}_{11} + \hat{\boldsymbol{y}}_2^{\top}\hat{\boldsymbol{\beta}}_{21}  
\end{align}
Note that $\hat{\boldsymbol{\beta}}_{22}$ is a square matrix (owing to $N_{x_2} = N_{y_2}$). For the discussion about the invertibility of $\hat{\boldsymbol{\beta}}_{22}$, see \cite{19}. 

{\color{black}
\noindent \textit{\textbf{Capabilities}}: Compared to the classic DPFL model $\boldsymbol{Y} = \boldsymbol{X}\boldsymbol{\beta}$, the model in \eqref{eq:rearrange} can handle bus-type variations. E.g., if a PV bus changes to a PQ bus, its reactive power injection moves from $\boldsymbol{Y}_2$ to $\boldsymbol{X}_1$ on the right-hand side of \eqref{eq:rearrange}, and its voltage becomes an unknown variable moving from $\boldsymbol{X}_2$ to $\boldsymbol{Y}_1$ on the left-hand side. The related coefficients adjust accordingly. The model remains valid without retraining because the learned coefficients still apply.

\noindent \textit{\textbf{Limitations}}: The scheme proposed in \cite{19} can lead to large linearization errors. This is because the invertibility of $\hat{\boldsymbol{\beta}}_{22}$ is not theoretically guaranteed. It can become near-singular if PV bus voltages are fixed in the training dataset or if active power injections at some nodes are consistently zero. This renders the resulting DPFL model unreliable. 

\noindent \textit{\textbf{Generalizability}}: This rearrange/bundle technique applies to most DPFL training algorithms discussed, except for the linearly constrained DPFL programming algorithm with coupling constraints. In this algorithm, predictors are confined to voltages/angles at the ends of a branch, and responses are the corresponding active/reactive branch flows, limiting the flexibility to rearrange or bundle them.

}



\subsection{Handling Data Multicollinearity}
In order to further mitigate the negative effects of multicollinearity, the following two techniques can be used to provide additional support.

\subsubsection{Increase Data Size}
The first technique is to increase the volume of the training dataset \cite{15,29}. Insufficient measurements may cause multicollinearity, and lengthening the columns in $\boldsymbol{X}$ can disrupt variable correlations. This effectiveness was observed in \cite{29} using real measurements from a three-phase distribution system in Switzerland, showing that a longer measurement time window significantly reduces singularity caused by multicollinearity.

\noindent \textit{\textbf{Capabilities}}: This technique is easy to implement and may reduce the influence of multicollinearity.

\noindent \textit{\textbf{Limitations}}: \textit{Firstly}, there is no theoretical guarantee that expanding the training dataset will consistently resolve multicollinearity. \textit{Secondly}, this method is ineffective without sufficient measurements.

\noindent \textit{\textbf{Generalizability}}: This technique is generally applicable to all previously discussed DPFL training algorithms when measurements are adequate.

\subsubsection{Replace Power Injection of Slack Bus}
The second technique applies to models like \eqref{eq:rearrange}. Here, the active power injection measurements at the slack bus are included in dataset $\boldsymbol{Y}_1$ \cite{19}. Including these measurements in $\boldsymbol{Y}_2$ may increase multicollinearity in matrix $[\boldsymbol{X}_{1} \  \boldsymbol{Y}_{2}]$ because the slack bus's active power injection approximates the sum of all PQ and PV buses' active power injections \cite{19} --- a highly collinear relationship \cite{19}. Therefore, the slack bus's active power injection can be separated from other active power injections.

\noindent \textit{\textbf{Capabilities}}: This technique is straightforward to implement and can mitigate multicollinearity.

\noindent \textit{\textbf{Limitations}}: \textit{Firstly}, separating the slack bus's active power injection does not fully resolve multicollinearity, as this issue also exists in voltage measurements. \textit{Secondly}, this technique has a narrow scope, only applicable when the bundling strategy for bus-type variation issues is used. Without this strategy, the slack bus's active power injection, usually unknown, does not appear with other known nodal active power injections.

\noindent \textit{\textbf{Generalizability}}: This technique is generally applicable to DPFL training algorithms when the bundling strategy for bus-type variation issues is employed.

\subsection{Handling Inherent Nonlinearity}\label{sec:Nonlinear}
In DPFL studies, there are a number of supportive techniques that can handle the nonlinearity inherent in the power flow equations. We categorize them into two classes: coordinate transformation \cite{1,4, 6, 9, 14, 18, 33, 27,28} and forgetting factors \cite{2,8,17,18}. 

\subsubsection{Coordinate Transformation} 
The transformation of variables' coordinates intends to map $\boldsymbol{x}$ and $\boldsymbol{y}$ onto new spaces where their relationship may enjoy a higher level of linearity. In DPFL, existing transformation techniques include voltage squaring \cite{4, 6, 9, 14, 18, 33}, voltage-angle coupling \cite{6}, and dimension lifting \cite{1}, as discussed below. 

\paragraph{Voltage Squaring}~\\

In AC power flow equations, voltages appear in quadratic forms like $v_i^2$ or $v_iv_j$. Thus, the AC model is more linear in the quadratic voltage space. Motivated by this, previous works attempt to linearize the AC model using voltage squares \cite{1_11, yang2017linearized, yang2018general}. \cite{yang2018general} tests different $n$ in $v^n$ for linearization accuracy, finding $n^* \approx 2$ has the lowest error. Consequently, voltage squares are often chosen as dependent variables in DPFL \cite{4, 6, 9, 14, 18, 33}. The coordinate transformation here is therefore described by:
\begin{align}
  \phi(\boldsymbol{y}) \coloneqq \  \{v_i\} \rightarrow \{\nu_i\}, \ \forall i
\end{align}
where $\nu_i = v_i^2$. Any DPFL training algorithm can then be used to learn $\boldsymbol{\beta}_{\phi}$ that suits $\boldsymbol{Y}_{\phi} = \boldsymbol{X}\boldsymbol{\beta}_{\phi}$, where $\boldsymbol{Y}_{\phi}$ consists of the measurements of the transformed dependent variables. 

\paragraph{Voltage-angle Coupling}~\\

The AC power flow equations in \eqref{eq:Pi} and \eqref{eq:Qi} can be equivalently converted into:
\begin{align}
  P_i & = G_{ii}\nu_i + \sum\nolimits_{j=1,j\neq i}^{N_b}\left(G_{ij}R_{ij} + B_{ij}C_{ij} \right)\notag \\
  Q_i & = -B_{ii}\nu_i + \sum\nolimits_{j=1,j\neq i}^{N_b}\left(G_{ij}C_{ij} - B_{ij}R_{ij} \right)\notag 
\end{align}
using the following coupled terms: 
\begin{align}
  \nu_i = v_i^2, \ R_{ij} =  v_iv_j\cos\theta_{ij}, \ C_{ij} =  v_iv_j\sin\theta_{ij} \notag
\end{align}
The above terms suggest that with the coordinate transformation defined as \cite{6}:
\begin{align}
  \phi(\boldsymbol{y}) \coloneqq \  \{v_i, v_j,\theta_{ij}\} \rightarrow \{\nu_i, R_{ij}, C_{ij}\}, \ \forall i,j, \label{eq:transform 2}
\end{align}
the AC equations amount to linear equations. In this case, the linear relationship $\boldsymbol{X} = \boldsymbol{Y}_{\phi} \boldsymbol{H}_{\phi}$ naturally holds, where $\boldsymbol{H}_{\phi}$ consists of the admittance parameters. It should be noted, however, that the non-square matrix $\boldsymbol{H}_{\phi}$ is not invertible. This means that one cannot derive the desired linear model $\boldsymbol{Y}_{\phi} = \boldsymbol{X}\boldsymbol{\beta}_{\phi}$ from $\boldsymbol{X} = \boldsymbol{Y}_{\phi} \boldsymbol{H}_{\phi}$. Consequently,  researchers employ DPFL training algorithms to directly estimate $\boldsymbol{\beta}_{\phi}$ in $\boldsymbol{Y}_{\phi} = \boldsymbol{X}\boldsymbol{\beta}_{\phi}$ \cite{6}. Importantly, \cite{6} shows that the resulting linear power flow model performs better than the DPFL model without using the transformation defined in \eqref{eq:transform 2}. 

\paragraph{Dimension Lifting}~\\

Mapping a nonlinear dynamic system to a higher-dimensional space where its evolution is more linear is a common problem. The Koopman operator theory \cite{1_22}, a widespread technique in nonlinear dynamics \cite{brunton2017koopman}, addresses this and has been applied in DPFL \cite{1}. 

Specifically, Koopman operator theory suggests transforming by lifting the dimension of the independent variable $\boldsymbol{x}$, i.e., 
\begin{align}
  \phi(\boldsymbol{x}) \coloneqq \  \{\boldsymbol{x}\} \rightarrow \left\{\left[ \boldsymbol{x}^{\top} \ \psi(\boldsymbol{x})^{\top} \right]^{\top} \right\} \label{eq:transform 4}
\end{align} 
where 
\begin{align}
  \psi(\boldsymbol{x}) = \left[\psi_1(\boldsymbol{x}) \ \cdots \ \psi_{N_l}(\boldsymbol{x})  \right]^{\top} \in \mathbb{R}^{N_l} \notag
\end{align}
Funtion $\psi_i(\cdot) \colon \mathbb{R}^{N_x \times 1} \to \mathbb{R}$, also known as the Koopman eigenfunction, pays particular attention to the $i$-th dimension of $\boldsymbol{x}$:
\begin{align} 
\psi_i(\boldsymbol{x}) =  f_{\text{lift}}(\boldsymbol{x} - \varsigma_i) 
\end{align}
where $\varsigma_i \in \mathbb{R}^{N_x \times 1}$ is a base vector used for lifting the $i$-th dimension of $\boldsymbol{x}$ \cite{1} or lifting the whole $\boldsymbol{x}$ \cite{korda2018linear}, which can be selected randomly within the interval of the variable range \cite{1, korda2018linear}. There are several classic selections for the dimension lifting function $f_{\text{lift}}(\cdot) \colon \mathbb{R}^{N_x \times 1} \to \mathbb{R}$, e.g., the logarithmic and exponential function families; see \cite{1, korda2018linear} for more details. After conducting the transformation in \eqref{eq:transform 4}, $\boldsymbol{\beta}_{\phi}$ can be computed using any DPFL algorithms such that $\boldsymbol{Y} = \boldsymbol{X}_{\phi}\boldsymbol{\beta}_{\phi}$ holds. 

{\color{black}
\noindent \textit{\textbf{Capabilities}}: \textit{Firstly}, these coordinate transformation techniques address the inherent nonlinearity of the AC power flow model, enhancing DPFL model performance. \textit{Secondly}, the voltage-squaring transformation is easy to implement and widely recognized. \textit{Thirdly}, the dimension-lifting transformation offers flexibility in variable selection and can be applied to original state variables or those already transformed. \textit{Lastly}, the dimension-lifting function is versatile, generally increasing linearity and reducing errors, as highlighted in the Koopman operator theory \cite{1}.

\noindent \textit{\textbf{Limitations}}: \textit{Firstly}, the voltage-angle-coupling transformation increases the number of variables significantly if branch variables are defined without specific knowledge of the power system's topology, leading to a larger training problem. \textit{Secondly}, the dimension-lifting transformation results in a model that represents the relationship between $\boldsymbol{y}$ and $\left[ \boldsymbol{x}^{\top} \ \psi(\boldsymbol{x})^{\top} \right]$, complicating its use in applications like optimal power flow or unit commitment due to added nonlinearity.

\noindent \textit{\textbf{Generalizability}}: \textit{Firstly}, none of these transformations are suitable for the linearly constrained DPFL programming algorithm with structure constraints, as they will lose the simplified symmetrical structure of the Jacobian matrix. \textit{Secondly}, the voltage-angle coupling transformation is incompatible with the linearly constrained DPFL programming algorithm that uses coupling constraints, as it removes the relationship between coefficients of $\theta_i$ and $\theta_j$.

}

\subsubsection{Forgetting Factor}\label{sec:local}
Compared to a global single model, using multiple local linear models better captures the nonlinearity of the AC model. Employing forgetting factors is a key technique in DPFL for training local models \cite{2,8,17,18}. This technique emphasizes recent observations by weighting them heavily and gradually lowering the weights of earlier observations \cite{2_33}. Forgetting factors are especially useful when the target system undergoes rapid changes in operating mode \cite{17}. Below, we examine two schemes for weighting forgetting factors: one for training DPFL models and the other for updating prior models.

\paragraph{Weighting Scheme for Training}~\\

We first generalize the objective function used in regression models (e.g., the least squares regression or ridge regression) as
\begin{align}
  \min_{\boldsymbol{\beta}} \quad  \sum_{i=1}^{N_s}\sum_{j=1}^{N_y} r_{ij}\left( y_{ij},\boldsymbol{x}_i, \boldsymbol{\beta}_j \right) \label{eq: quad_obj}
\end{align}
where $r_{ij}(\cdot)$ is a general description of the residual, e.g., 
\begin{align} 
r_{ij}\left( y_{ij},\boldsymbol{x}_i, \boldsymbol{\beta}_j \right) = \Vert y_{ij}-\boldsymbol{x}_i^{\top}\boldsymbol{\beta}_j \Vert_2^2
\end{align}
if the ordinary least squares regression is used. In addition, to distinguish between earlier and recent observations, the observations in $\boldsymbol{X}$ and $\boldsymbol{Y}$ are ordered chronologically, e.g., $\boldsymbol{x}_i$ is older than $\boldsymbol{x}_{i+1}$. To support training DPFL models, forgetting factors can be integrated as follows \cite{18}:
\begin{align}
  \min_{\boldsymbol{\beta}} \quad  \sum_{i=1}^{N_s} \varpi^{N_s-i} \sum_{j=1}^{N_y} r_{ij}\left( y_{ij},\boldsymbol{x}_i, \boldsymbol{\beta}_j \right) \label{eq: quad_obj_forget}
\end{align}
where $\varpi \in \mathbb{R}$ is some pre-defined constant in the range $ (0, \ 1]$. Clearly, the above weighting scheme places heavier weights on more recent observations in the objective function given in \eqref{eq: quad_obj_forget}. This objective function can then be employed to train linear models using most DPFL training algorithms. 

\paragraph{Weighting Scheme for Updating}~\\

For demonstration purposes, we use the recursive partial least squares regression model. This model is designed to incrementally update DPFL models. Specifically, instead of using the original rule, i.e.,
\begin{align}
  \widetilde{\boldsymbol{X}}[t+1] = \left[ \begin{array}{c}
    \boldsymbol{C}[t]^{\top} \\
    \boldsymbol{x}_{t+1} \\
    \end{array} \right],\ \widetilde{\boldsymbol{Y}}[t+1] = \left[ \begin{array}{c}
      \boldsymbol{\varGamma}[t] \boldsymbol{R}[t]^{\top} \\
      \boldsymbol{y}_{t+1} \\
      \end{array} \right]  \notag
\end{align}
to update the growing datasets $\widetilde{\boldsymbol{X}}[t+1]$ and $\widetilde{\boldsymbol{Y}}[t+1]$, the following scheme with forgetting factors is utilized \cite{17}:
\begin{align}
  \widetilde{\boldsymbol{X}}[t+1] \!=\! \left[ \begin{array}{c}
    \varpi \boldsymbol{C}[t]^{\top} \\
    \boldsymbol{x}_{t+1} \\
    \end{array} \right], \widetilde{\boldsymbol{Y}}[t+1] \!=\! \left[ \begin{array}{c}
      \varpi \boldsymbol{\varGamma}[t] \boldsymbol{R}[t]^{\top} \\
      \boldsymbol{y}_{t+1} \\
      \end{array} \right]  \label{eq:RPLS_decom_forget}
\end{align}
where $\varpi$ again belongs to the range $ (0, \ 1]$. According to \cite{17}, $\varpi$ can be set via cross-validation. After the integration of $\varpi$, the contributions from earlier measurements become lesser during the incremental update. 

{\color{black}
\noindent \textit{\textbf{Capabilities}}: These two schemes are easy to implement and can track the current operating state, helping identify a suitable linear model within the nonlinear AC manifold.

\noindent \textit{\textbf{Limitations}}: \textit{Firstly}, using a forgetting factor adds a hyperparameter that needs tuning, increasing computational burden. \textit{Secondly}, if $\varpi$ is less than 1, $\varpi^{N_s-i}$ can become very small even for large $i$. E.g., with $\varpi = 0.6$, $N_s=300$, and $i=280$, $\varpi^{N_s-i}$ drops to $3\times 10^{-5}$, making earlier samples, from $i=1$ to $i=280$, actually excluded from training. This raises concerns about having enough samples to train a DPFL model. This issue also applies to the updating scheme, as $\varpi$ is cumulatively multiplied as well.

\noindent \textit{\textbf{Generalizability}}: \textit{Firstly}, the training scheme is broadly applicable to all tailored DPFL algorithms and those related to support vector regression. \textit{Secondly}, the updating scheme is specific to the recursive partial least squares regression.

}

\subsection{Integration of Physical Knowledge}
Integrating accessible physical knowledge is a potential way to improve the training performance of DPFL models. So far, researchers have exploited two types of physical knowledge in DPFL, namely (i) the grid topology \cite{33,9} and (ii) physical depedencies between system variables by integrating physics-driven power flow linearization (PPFL) models \cite{18, 10, 23}, as discussed below. 

\subsubsection{Grid Topology}
The grid topology shows how lines and buses are connected, helping locate related variables. In standard DPFL models, a line flow is described by all nodal variables. With grid topology knowledge, line flow can be described using only the variables of its two terminals. Thus, DPFL model training splits into subproblems, each aiming to train a linear relationship between a line flow (active or reactive) and its terminals \cite{33,9}, e.g.,
\begin{align} 
P_{ij} \stackrel{\text{Linear}}{\longleftarrow\! \longrightarrow}
 \{v_i, v_j, \theta_i, \theta_j, P_i, Q_i \}
\end{align}
Any DPFL training algorithm can then be applied to the aforementioned subproblem.

\subsubsection{PPFL Model}
A PPFL model integrates three types of physical knowledge: (i) physical dependencies between the variables using a model thereof, (ii) the grid topology, and (iii) physical parameters. Based on which types of knowledge are used in DPFL training, the integration techniques can be grouped into two categories: (i) coefficient optimization \cite{23} and (ii) error correction \cite{9960825, 10}.

\paragraph{Coefficient Optimization}~\\

Impedance parameters in a PPFL model serve as coefficients mapping independent to dependent variables. Even if known, they can be optimized in a data-driven way to improve linearization accuracy \cite{23, 9960825}. This integrates PPFL models into the DPFL training process, absorbing all three types of physical knowledge. 

Let us take the classic DC power flow model as an example:
\begin{align}
  P_{ij} =  \theta_{ij}/z_{x,ij}, \ \forall i,j 
\end{align}
where $z_{x,ij}$ denotes the series reactance of line $i-j$. To further optimize the coefficient in order to improve the linearization accuracy, an additional coefficient $\beta_{ij}$ can be introduced, altering the above PPFL model to \cite{23}:
\begin{align}
  P_{ij} =  \beta_{ij} \theta_{ij}/z_{x,ij}, \ \forall i,j 
\end{align}
With the datasets of $P_{ij}$ and $\theta_{ij}$, $\beta_{ij}$ can be optimized using any DPFL training algorithm reviewed before. A similar approach is discussed in our previous work \cite{9960825}, where the state-independent PPFL model presented in \cite{11_14} is utilized for data-driven parameter optimization. Results show that the data-driven parameter optimization can significantly reduce the linearization error for both voltage and angle, by one to two orders of magnitude \cite{9960825}.

\paragraph{Error Correction}~\\

Linearization errors are common in PPFL models. These errors can be reduced by modeling them as linear functions of independent variables using any DPFL training algorithm. The error models are then added as corrections to the PPFL model while keeping it linear. This error correction process is another way to integrate PPFL models into DPFL training when impedance values are available \cite{9960825, 10}.

Specifically, let us denote the AC power flow model as
\begin{align}
  \boldsymbol{y}^{AC} & = \boldsymbol{y}^{PHY} + \Delta\boldsymbol{y}  
\end{align}
where 
\begin{align} 
\boldsymbol{y}^{PHY} =  \boldsymbol{A}^{\top}\boldsymbol{x} \end{align} 
is a PPFL model, and $\Delta\boldsymbol{y}$ refers to the approximation error. We further define the dataset of $\Delta\boldsymbol{y}$ as $\Delta\boldsymbol{Y}$. The coefficient $\boldsymbol{\beta}^{ERR}$ in 
\begin{align} 
\Delta\boldsymbol{Y} = \boldsymbol{X}\boldsymbol{\beta}^{ERR}
\end{align}
is found using any DPFL training algorithm. The resulting $\boldsymbol{\beta}^{ERR}$, i.e., $\hat{\boldsymbol{\beta}}^{ERR}$, is then be integrated into  the original PPFL model, yielding the corrected linear power flow model: 
\begin{align} 
\boldsymbol{y}^{COR} = \left[\boldsymbol{A} + \hat{\boldsymbol{\beta}}^{ERR}\right]^{\top }\boldsymbol{x}
\end{align}

{\color{black}
\noindent \textit{\textbf{Capabilities}}: Integrating physical knowledge into the DPFL training process can enhance model accuracy and compensate for limited training data.

\noindent \textit{\textbf{Limitations}}: These techniques share the limitations of linearly constrained DPFL programming algorithms (as detailed in Section II-B-1), including information loss from restricted variables, incompatibility with normalized data, inaccuracies in the physical knowledge, and limited model applicability.

\noindent \textit{\textbf{Generalizability}}: \textit{Firstly}, incorporating grid topology is feasible for most DPFL training algorithms. \textit{Secondly}, integrating PPFL models is incompatible with linearly constrained DPFL programming algorithms, as the original physical meanings needed for computing the Jacobian matrix or coupling relationships will be lost. 
}

\section{Conclusion}

{\color{black}
This first part of the two-part tutorial provides a comprehensive reexamination and critical analysis of DPFL theories. This paper has systematically separated DPFL supportive techniques from DPFL training algorithms. It then thoroughly revisited and discussed the mathematical derivations and solutions of all DPFL training algorithms and supportive techniques, providing a detailed, methodological guide for researchers. This contribution bridges the gaps in the literature, including the lack of a unified mathematical framework for all DPFL methods, the absence of a comprehensive discussion on their capabilities and limitations, and the insufficient discussion of the applicability of DPFL supportive techniques. Any method has its strengths and weaknesses. The rigorous examination done in this paper intends to provide a thorough and balanced assessment, thereby offering valuable insights for potential future enhancements.

Furthermore, this paper observes that many DPFL training algorithms and supportive techniques share similar capabilities and limitations. This highlights a limitation inherent in theoretical analysis: when methods bear theoretical similarities, predicting their actual performance differences becomes challenging. Therefore, a comprehensive numerical evaluation of all DPFL methods across multiple aspects becomes essential, which is lacking in the current literature. The second part \cite{partII} of the tutorial aims to bridge this gap, mainly focusing on deepening the field's understanding of all the DPFL methods' performance in practical scenarios.

}



\bibliographystyle{elsarticle-num} 
\bibliography{cas-refs}

\begin{thebibliography}{10}
\expandafter\ifx\csname url\endcsname\relax
  \def\url#1{\texttt{#1}}\fi
\expandafter\ifx\csname urlprefix\endcsname\relax\def\urlprefix{URL }\fi
\expandafter\ifx\csname href\endcsname\relax
  \def\href#1#2{#2} \def\path#1{#1}\fi

\bibitem{9914682}
Y.~Chen, F.~Pan, F.~Qiu, A.~S. Xavier, T.~Zheng, M.~Marwali, B.~Knueven, Y.~Guan, P.~B. Luh, L.~Wu, B.~Yan, M.~A. Bragin, H.~Zhong, A.~Giacomoni, R.~Baldick, B.~Gisin, Q.~Gu, R.~Philbrick, F.~Li, Security-constrained unit commitment for electricity market: Modeling, solution methods, and future challenges, IEEE Transactions on Power Systems 38~(5) (2023) 4668--4681.
\newblock \href {https://doi.org/10.1109/TPWRS.2022.3213001} {\path{doi:10.1109/TPWRS.2022.3213001}}.

\bibitem{molzahn2019survey}
D.~K. Molzahn, I.~A. Hiskens, et~al., A survey of relaxations and approximations of the power flow equations, Foundations and Trends{\textregistered} in Electric Energy Systems 4~(1-2) (2019) 1--221.

\bibitem{23}
X.~Li, K.~Hedman, Data driven linearized ac power flow model with regression analysis, arXiv preprint arXiv:1811.09727 (2018).

\bibitem{8_1}
Z.~Li, J.~Yu, Q.~Wu, Approximate linear power flow using logarithmic transform of voltage magnitudes with reactive power and transmission loss consideration, IEEE Transactions on Power Systems 33~(4) (2017) 4593--4603.

\bibitem{dorfler2023data}
F.~D{\"o}rfler, Data-driven control: Part one of two: A special issue sampling from a vast and dynamic landscape, IEEE Control Systems Magazine 43~(5) (2023) 24--27.

\bibitem{Powertech}
M.~Jia, G.~Hug, Overview of data-driven power flow linearization, in: 2023 IEEE Belgrade PowerTech, 2023, pp. 01--06.
\newblock \href {https://doi.org/10.1109/PowerTech55446.2023.10202779} {\path{doi:10.1109/PowerTech55446.2023.10202779}}.

\bibitem{19}
Y.~Liu, N.~Zhang, Y.~Wang, J.~Yang, C.~Kang, Data-driven power flow linearization: A regression approach, IEEE Transactions on Smart Grid 10~(3) (2018) 2569--2580.

\bibitem{14}
J.~Chen, W.~Li, W.~Wu, T.~Zhu, Z.~Wang, C.~Zhao, Robust data-driven linearization for distribution three-phase power flow, in: 2020 IEEE 4th Conference on Energy Internet and Energy System Integration (EI2), IEEE, 2020, pp. 1527--1532.

\bibitem{14_16}
S.~Gajare, A.~K. Pradhan, V.~Terzija, A method for accurate parameter estimation of series compensated transmission lines using synchronized data, IEEE Transactions on Power Systems 32~(6) (2017) 4843--4850.

\bibitem{14_18}
P.~Huynh, H.~Zhu, Q.~Chen, A.~E. Elbanna, Data-driven estimation of frequency response from ambient synchrophasor measurements, IEEE Transactions on Power Systems 33~(6) (2018) 6590--6599.

\bibitem{26_14}
A.~Von~Meier, E.~Stewart, A.~McEachern, M.~Andersen, L.~Mehrmanesh, Precision micro-synchrophasors for distribution systems: A summary of applications, IEEE Transactions on Smart Grid 8~(6) (2017) 2926--2936.

\bibitem{30}
J.~Zhang, P.~Wang, N.~Zhang, Distribution network admittance matrix estimation with linear regression, IEEE Transactions on Power Systems 36~(5) (2021) 4896--4899.

\bibitem{4_22}
P.~Huynh, H.~Zhu, Q.~Chen, A.~E. Elbanna, Data-driven estimation of frequency response from ambient synchrophasor measurements, IEEE Transactions on Power Systems 33~(6) (2018) 6590--6599.

\bibitem{terzija2010wide}
V.~Terzija, G.~Valverde, D.~Cai, P.~Regulski, V.~Madani, J.~Fitch, S.~Skok, M.~M. Begovic, A.~Phadke, Wide-area monitoring, protection, and control of future electric power networks, Proceedings of the IEEE 99~(1) (2010) 80--93.

\bibitem{de2010synchronized}
J.~De~La~Ree, V.~Centeno, J.~S. Thorp, A.~G. Phadke, Synchronized phasor measurement applications in power systems, IEEE Transactions on smart grid 1~(1) (2010) 20--27.

\bibitem{2}
Y.~Liu, Z.~Li, Y.~Zhou, A physics-based and data-driven linear three-phase power flow model for distribution power systems, arXiv preprint arXiv:2103.10147 (2021).

\bibitem{11}
Y.~Liu, Y.~Wang, N.~Zhang, D.~Lu, C.~Kang, A data-driven approach to linearize power flow equations considering measurement noise, IEEE Transactions on Smart Grid 11~(3) (2019) 2576--2587.

\bibitem{11_14}
J.~Yang, N.~Zhang, C.~Kang, Q.~Xia, A state-independent linear power flow model with accurate estimation of voltage magnitude, IEEE Transactions on Power Systems 32~(5) (2016) 3607--3617.

\bibitem{7}
Z.~Shao, Q.~Zhai, J.~Wu, X.~Guan, Data based linear power flow model: Investigation of a least-squares based approximation, IEEE Transactions on Power Systems 36~(5) (2021) 4246--4258.

\bibitem{11_13}
S.~M. Fatemi, S.~Abedi, G.~Gharehpetian, S.~H. Hosseinian, M.~Abedi, Introducing a novel dc power flow method with reactive power considerations, IEEE Transactions on Power Systems 30~(6) (2014) 3012--3023.

\bibitem{wang2017linear}
Y.~Wang, N.~Zhang, H.~Li, J.~Yang, C.~Kang, Linear three-phase power flow for unbalanced active distribution networks with pv nodes, CSEE Journal of Power and Energy Systems 3~(3) (2017) 321--324.

\bibitem{yang2016state}
J.~Yang, N.~Zhang, C.~Kang, Q.~Xia, A state-independent linear power flow model with accurate estimation of voltage magnitude, IEEE Transactions on Power Systems 32~(5) (2016) 3607--3617.

\bibitem{fan2021error}
Z.~Fan, Z.~Yang, J.~Yu, Error bound restriction of linear power flow model, IEEE Transactions on Power Systems 37~(1) (2021) 808--811.

\bibitem{1}
L.~Guo, Y.~Zhang, X.~Li, Z.~Wang, Y.~Liu, L.~Bai, C.~Wang, Data-driven power flow calculation method: A lifting dimension linear regression approach, IEEE Transactions on Power Systems (2021).

\bibitem{29_11}
S.~Frank, J.~Sexauer, S.~Mohagheghi, Temperature-dependent power flow, IEEE Transactions on Power Systems 28~(4) (2013) 4007--4018.

\bibitem{27}
J.~Yu, Y.~Weng, R.~Rajagopal, Robust mapping rule estimation for power flow analysis in distribution grids, in: 2017 North American Power Symposium (NAPS), IEEE, 2017, pp. 1--6.

\bibitem{4}
J.~Chen, W.~Wu, L.~A. Roald, Data-driven piecewise linearization for distribution three-phase stochastic power flow, IEEE Transactions on Smart Grid (2021).

\bibitem{9027950}
J.~Zhang, Y.~Wang, Y.~Weng, N.~Zhang, Topology identification and line parameter estimation for non-pmu distribution network: A numerical method, IEEE Transactions on Smart Grid 11~(5) (2020) 4440--4453.
\newblock \href {https://doi.org/10.1109/TSG.2020.2979368} {\path{doi:10.1109/TSG.2020.2979368}}.

\bibitem{5_10}
Y.~Weng, Y.~Liao, R.~Rajagopal, Distributed energy resources topology identification via graphical modeling, IEEE Transactions on Power Systems 32~(4) (2016) 2682--2694.

\bibitem{5_11}
Y.~Wang, K.~Tan, X.~Y. Peng, P.~L. So, Coordinated control of distributed energy-storage systems for voltage regulation in distribution networks, IEEE transactions on power delivery 31~(3) (2015) 1132--1141.

\bibitem{19_18}
D.~Zhang, J.~Li, D.~Hui, Coordinated control for voltage regulation of distribution network voltage regulation by distributed energy storage systems, Protection and Control of Modern Power Systems 3~(1) (2018) 1--8.

\bibitem{29_8}
A.~Al-Othman, M.~Irving, Analysis of confidence bounds in power system state estimation with uncertainty in both measurements and parameters, Electric power systems research 76~(12) (2006) 1011--1018.

\bibitem{29_9}
D.-H. Choi, L.~Xie, Impact analysis of locational marginal price subject to power system topology errors, in: 2013 IEEE International Conference on Smart Grid Communications (SmartGridComm), IEEE, 2013, pp. 55--60.

\bibitem{29_10}
K.~Clements, P.~Davis, Detection and identification of topology errors in electric power systems, IEEE Transactions on Power systems 3~(4) (1988) 1748--1753.

\bibitem{19_19}
S.~Gajare, A.~K. Pradhan, V.~Terzija, A method for accurate parameter estimation of series compensated transmission lines using synchronized data, IEEE Transactions on Power Systems 32~(6) (2017) 4843--4850.

\bibitem{27_11}
M.~Fan, V.~Vittal, G.~T. Heydt, R.~Ayyanar, Probabilistic power flow studies for transmission systems with photovoltaic generation using cumulants, IEEE Transactions on Power Systems 27~(4) (2012) 2251--2261.

\bibitem{20}
C.~Qin, L.~Wang, Z.~Han, J.~Zhao, W.~Wang, A modified data-driven regression model for power flow analysis, in: 2019 IEEE 8th Data Driven Control and Learning Systems Conference (DDCLS), IEEE, 2019, pp. 794--799.

\bibitem{7_9}
M.~Z. Kamh, R.~Iravani, A sequence frame-based distributed slack bus model for energy management of active distribution networks, IEEE Transactions on smart Grid 3~(2) (2012) 828--836.

\bibitem{9_15}
T.~Carriere, G.~Kariniotakis, An integrated approach for value-oriented energy forecasting and data-driven decision-making application to renewable energy trading, IEEE transactions on smart grid 10~(6) (2019) 6933--6944.

\bibitem{16}
X.~Li, Fast heuristic ac power flow analysis with data-driven enhanced linearized model, Energies 13~(13) (2020) 3308.

\bibitem{18}
H.~Xu, A.~D. Dom{\'\i}nguez-Garc{\'\i}a, V.~V. Veeravalli, P.~W. Sauer, Data-driven voltage regulation in radial power distribution systems, IEEE Transactions on Power Systems 35~(3) (2019) 2133--2143.

\bibitem{13}
S.~hentong, Z.~Qiaozhu, W.~Jiang, G.~Xiaohong, Data based linearization: Least-squares based approximation, arXiv preprint arXiv:2007.02494 (2020).

\bibitem{34}
P.~Li, W.~Wu, X.~Wan, B.~Xu, A data-driven linear optimal power flow model for distribution networks, IEEE Transactions on Power Systems (2022).

\bibitem{8}
Y.~Liu, Z.~Li, Y.~Zhou, Data-driven-aided linear three-phase power flow model for distribution power systems, IEEE Transactions on Power Systems (2021).

\bibitem{29}
C.~Mugnier, K.~Christakou, J.~Jaton, M.~De~Vivo, M.~Carpita, M.~Paolone, Model-less/measurement-based computation of voltage sensitivities in unbalanced electrical distribution networks, in: 2016 Power Systems Computation Conference (PSCC), IEEE, 2016, pp. 1--7.

\bibitem{12}
S.~Powell, A.~Ivanova, D.~Chassin, Fast solutions in power system simulation through coupling with data-driven power flow models for voltage estimation, arXiv preprint arXiv:2001.01714 (2020).

\bibitem{10}
Y.~Tan, Y.~Chen, Y.~Li, Y.~Cao, Linearizing power flow model: A hybrid physical model-driven and data-driven approach, IEEE Transactions on Power Systems 35~(3) (2020) 2475--2478.

\bibitem{17}
S.~Nowak, Y.~C. Chen, L.~Wang, Measurement-based optimal der dispatch with a recursively estimated sensitivity model, IEEE Transactions on Power Systems 35~(6) (2020) 4792--4802.

\bibitem{6}
Y.~Chen, C.~Wu, J.~Qi, Data-driven power flow method based on exact linear regression equations, Journal of Modern Power Systems and Clean Energy (2021).

\bibitem{24}
J.~Zhang, Z.~Wang, X.~Zheng, L.~Guan, C.~Chung, Locally weighted ridge regression for power system online sensitivity identification considering data collinearity, IEEE Transactions on Power Systems 33~(2) (2017) 1624--1634.

\bibitem{28}
J.~Yu, Y.~Weng, R.~Rajagopal, Mapping rule estimation for power flow analysis in distribution grids, arXiv preprint arXiv:1702.07948 (2017).

\bibitem{liu2023data}
Y.~Liu, Z.~Li, S.~Sun, A data-driven method for online constructing linear power flow model, IEEE Transactions on Industry Applications (2023).

\bibitem{9}
Y.~Liu, B.~Xu, A.~Botterud, N.~Zhang, C.~Kang, Bounding regression errors in data-driven power grid steady-state models, IEEE Transactions on Power Systems 36~(2) (2020) 1023--1033.

\bibitem{33}
Z.~Shao, Q.~Zhai, Z.~Han, X.~Guan, A linear ac unit commitment formulation: An application of data-driven linear power flow model, International Journal of Electrical Power \& Energy Systems 145 (2023) 108673.

\bibitem{3}
Y.~Liu, Z.~Li, J.~Zhao, Robust data-driven linear power flow model with probability constrained worst-case errors, arXiv preprint arXiv:2112.10320 (2021).

\bibitem{shao2023physical}
Z.~Shao, Q.~Zhai, X.~Guan, Physical-model-aided data-driven linear power flow model: an approach to address missing training data, IEEE Transactions on Power Systems (2023).

\bibitem{partII}
M.~Jia, G.~Hug, N.~Zhang, Z.~Wang, Y.~Wang, C.~Kang, Data-driven power flow linearization - part ii: Simulation (2024).

\bibitem{24_6}
J.~Zhang, L.~Guan, C.~Chung, Instantaneous sensitivity identification in power systems-challenges and technique roadmap, in: 2016 IEEE Power and Energy Society General Meeting (PESGM), IEEE, 2016, pp. 1--5.

\bibitem{26}
P.~Li, H.~Su, C.~Wang, Z.~Liu, J.~Wu, Pmu-based estimation of voltage-to-power sensitivity for distribution networks considering the sparsity of jacobian matrix, IEEE Access 6 (2018) 31307--31316.

\bibitem{7_28}
B.~Philippe, A.~H. Sameh, D.~Padua, Linear least squares and orthogonal factorization. (2011).

\bibitem{8_35}
E.~Kyriakides, S.~Suryanarayanan, G.~T. Heydt, State estimation in power engineering using the huber robust regression technique, IEEE Transactions on Power Systems 20~(2) (2005) 1183--1184.

\bibitem{8_34}
S.~Boyd, S.~P. Boyd, L.~Vandenberghe, Convex optimization, Cambridge university press, 2004.

\bibitem{davidson2004econometric}
R.~Davidson, J.~G. MacKinnon, et~al., Econometric theory and methods, Vol.~5, Oxford University Press New York, 2004.

\bibitem{overview}
I.~Markovsky, S.~Van~Huffel, Overview of total least-squares methods, Signal processing 87~(10) (2007) 2283--2302.

\bibitem{7084175}
M.~Badoni, A.~Singh, B.~Singh, Variable forgetting factor recursive least square control algorithm for dstatcom, IEEE Transactions on Power Delivery 30~(5) (2015) 2353--2361.
\newblock \href {https://doi.org/10.1109/TPWRD.2015.2422139} {\path{doi:10.1109/TPWRD.2015.2422139}}.

\bibitem{5}
G.~Jain, S.~Sidar, D.~Kiran, Alternative regression approach for data-driven power flow linearization methods, in: 2021 9th IEEE International Conference on Power Systems (ICPS), IEEE, 2021, pp. 1--6.

\bibitem{wold1984collinearity}
S.~Wold, A.~Ruhe, H.~Wold, W.~Dunn, Iii, The collinearity problem in linear regression. the partial least squares (pls) approach to generalized inverses, SIAM Journal on Scientific and Statistical Computing 5~(3) (1984) 735--743.

\bibitem{de1993simpls}
S.~De~Jong, Simpls: an alternative approach to partial least squares regression, Chemometrics and intelligent laboratory systems 18~(3) (1993) 251--263.

\bibitem{19_26}
R.~Rosipal, N.~Kr{\"a}mer, Overview and recent advances in partial least squares, in: International Statistical and Optimization Perspectives Workshop" Subspace, Latent Structure and Feature Selection", Springer, 2005, pp. 34--51.

\bibitem{qin1993partial}
S.~J. Qin, Partial least squares regression for recursive system identification, in: Proceedings of 32nd IEEE Conference on Decision and Control, IEEE, 1993, pp. 2617--2622.

\bibitem{wold1975path}
H.~Wold, Path models with latent variables: The nipals approach, in: Quantitative sociology, Elsevier, 1975, pp. 307--357.

\bibitem{alin2009comparison}
A.~Alin, Comparison of pls algorithms when number of objects is much larger than number of variables, Statistical papers 50~(4) (2009) 711--720.

\bibitem{17_30}
S.~J. Qin, Recursive pls algorithms for adaptive data modeling, Computers \& Chemical Engineering 22~(4-5) (1998) 503--514.

\bibitem{hoerl1970ridge}
A.~E. Hoerl, R.~W. Kennard, Ridge regression: Biased estimation for nonorthogonal problems, Technometrics 12~(1) (1970) 55--67.

\bibitem{24_18}
A.~E. Hoerl, R.~W. Kannard, K.~F. Baldwin, Ridge regression: some simulations, Communications in Statistics-Theory and Methods 4~(2) (1975) 105--123.

\bibitem{24_5}
J.~Zhang, C.~Y. Chung, Y.~Han, Online damping ratio prediction using locally weighted linear regression, IEEE Transactions on Power Systems 31~(3) (2016) 1954--1962.
\newblock \href {https://doi.org/10.1109/TPWRS.2015.2448104} {\path{doi:10.1109/TPWRS.2015.2448104}}.

\bibitem{4_35}
P.~S. Bradley, O.~L. Mangasarian, K-plane clustering, Journal of Global optimization 16~(1) (2000) 23--32.

\bibitem{4_26}
N.~Manwani, P.~Sastry, K-plane regression, Information Sciences 292 (2015) 39--56.

\bibitem{14_29}
A.~J. Smola, B.~Sch{\"o}lkopf, A tutorial on support vector regression, Statistics and computing 14~(3) (2004) 199--222.

\bibitem{flake2002efficient}
G.~W. Flake, S.~Lawrence, Efficient svm regression training with smo, Machine Learning 46~(1) (2002) 271--290.

\bibitem{27_15}
B.~Sch{\"o}lkopf, A.~J. Smola, F.~Bach, et~al., Learning with kernels: support vector machines, regularization, optimization, and beyond, MIT press, 2002.

\bibitem{25}
J.~Yu, W.~Dai, W.~Li, X.~Liu, J.~Liu, Optimal reactive power flow of interconnected power system based on static equivalent method using border pmu measurements, IEEE Transactions on Power Systems 33~(1) (2017) 421--429.

\bibitem{3_10}
W.~Wei, Tutorials on advanced optimization methods, arXiv preprint arXiv:2007.13545 (2020).

\bibitem{3_9}
Y.~Zhang, S.~Shen, J.~L. Mathieu, Distributionally robust chance-constrained optimal power flow with uncertain renewables and uncertain reserves provided by loads, IEEE Transactions on Power Systems 32~(2) (2016) 1378--1388.

\bibitem{jiang2016data}
R.~Jiang, Y.~Guan, Data-driven chance constrained stochastic program, Mathematical Programming 158~(1) (2016) 291--327.

\bibitem{15}
R.~Hu, Q.~Li, S.~Lei, Ensemble learning based linear power flow, in: 2020 IEEE Power \& Energy Society General Meeting (PESGM), IEEE, 2020, pp. 1--5.

\bibitem{1_11}
J.~Lavaei, S.~H. Low, Zero duality gap in optimal power flow problem, IEEE Transactions on Power systems 27~(1) (2011) 92--107.

\bibitem{yang2017linearized}
Z.~Yang, H.~Zhong, A.~Bose, T.~Zheng, Q.~Xia, C.~Kang, A linearized opf model with reactive power and voltage magnitude: A pathway to improve the mw-only dc opf, IEEE Transactions on Power Systems 33~(2) (2017) 1734--1745.

\bibitem{yang2018general}
Z.~Yang, K.~Xie, J.~Yu, H.~Zhong, N.~Zhang, Q.~Xia, A general formulation of linear power flow models: Basic theory and error analysis, IEEE Transactions on Power Systems 34~(2) (2018) 1315--1324.

\bibitem{1_22}
B.~O. Koopman, Hamiltonian systems and transformation in hilbert space, Proceedings of the National Academy of Sciences 17~(5) (1931) 315--318.

\bibitem{brunton2017koopman}
S.~Brunton, E.~Kaiser, N.~Kutz, Koopman operator theory: Past, present, and future, in: APS Division of Fluid Dynamics Meeting Abstracts, 2017, pp. L27--004.

\bibitem{korda2018linear}
M.~Korda, I.~Mezi{\'c}, Linear predictors for nonlinear dynamical systems: Koopman operator meets model predictive control, Automatica 93 (2018) 149--160.

\bibitem{2_33}
M.~Beza, M.~Bongiorno, Application of recursive least squares algorithm with variable forgetting factor for frequency component estimation in a generic input signal, IEEE Transactions on Industry Applications 50~(2) (2013) 1168--1176.

\bibitem{9960825}
M.~Jia, Q.~Cao, C.~Shen, G.~Hug, Frequency-control-aware probabilistic load flow: An analytical method, IEEE Transactions on Power Systems (2022) 1--16\href {https://doi.org/10.1109/TPWRS.2022.3223884} {\path{doi:10.1109/TPWRS.2022.3223884}}.

\end{thebibliography}




\end{document}